\documentclass[10pt,journal,compsoc]{IEEEtran}
\usepackage[utf8]{inputenc} 
\usepackage[T1]{fontenc}    
\usepackage{hyperref}       
\usepackage{url}            
\usepackage{booktabs}       
\usepackage{amsfonts}       
\usepackage{nicefrac}       
\usepackage{microtype}      
\usepackage{xcolor}         
\usepackage{graphicx}
\usepackage{booktabs}
\usepackage{multirow}
\usepackage{color}

\usepackage{graphicx}
\usepackage{comment}
\usepackage{amsmath}
\usepackage{amsthm}
\usepackage{amssymb}
\usepackage{array}
\usepackage{float}
\usepackage{subfig}
\usepackage{multirow}
\usepackage{longtable}
\usepackage{wrapfig, epsfig, epstopdf}
\usepackage{algorithm}
\usepackage{diagbox}
\usepackage[dash,dot]{dashundergaps}
\usepackage{algpseudocode}
\usepackage{tikz}
\usepackage{comment}
\usepackage{multirow}
\usepackage[11pt]{moresize}
\hypersetup{colorlinks=true,citecolor=black,linkcolor=black}
\usepackage[makeroom]{cancel}
\usepackage{stmaryrd}
\usepackage{booktabs, makecell}
\usepackage{makecell}
\usepackage{pifont}
\usepackage{lipsum}
\usepackage{mathtools}

\usepackage{lipsum}
\usepackage{bm}
\usepackage{lineno}
\usepackage{diagbox}
\usepackage{multirow}
\usepackage{caption}
\usepackage{bbm}
\usepackage{xfrac}
\usepackage{wrapfig,lipsum,booktabs}
\usepackage[hashEnumerators,smartEllipses]{markdown}

\definecolor{b2}{RGB}{51,153,255}
\newtheorem{theorem}{Theorem}[section]
\newtheorem{lemma}[theorem]{Lemma}
\newtheorem{definition}[theorem]{Definition}

\newtheorem{proposition}[theorem]{Proposition}
\newtheorem{corollary}[theorem]{Corollary}

\newtheorem{assumption}[theorem]{Assumption}

\renewcommand{\hat}{\widehat}

\newcommand{\N}{\mathcal{N}}

\renewcommand{\b}{\mathbf}
\renewcommand{\bm}{\mathbbm}

\DeclareMathOperator{\A}{{\mathcal A}}

\DeclareMathOperator{\D}{{\mathcal D}}

\DeclareMathOperator{\X}{{\mathcal X}}

\DeclareMathOperator{\Y}{{\mathcal Y}}
\DeclareMathOperator{\cH}{{\mathcal H}}
\DeclareMathOperator{\z}{{\mathbf z}}
\DeclareMathOperator{\x}{{\mathbf x}}

\newcommand{\cmark}{\ding{51}}%
\newcommand{\plus}{\ding{59}}%

\usepackage{amsmath,amsfonts,bm}

\def\eqref#1{equation~\ref{#1}}

\def\1{\bm{1}}

\DeclareMathAlphabet{\mathsfit}{\encodingdefault}{\sfdefault}{m}{sl}
\SetMathAlphabet{\mathsfit}{bold}{\encodingdefault}{\sfdefault}{bx}{n}

\newcommand{\E}{\mathbb{E}}

\newcommand{\R}{\mathbb{R}}

\DeclareMathOperator*{\argmin}{arg\,min}

\newcommand{\independent}{\perp\mkern-9.5mu\perp}

\makeatletter
\newcommand*{\da@rightarrow}{\mathchar"0\hexnumber@\symAMSa 4B }
\newcommand*{\da@leftarrow}{\mathchar"0\hexnumber@\symAMSa 4C }
\newcommand*{\xdashrightarrow}[2][]{%
  \mathrel{%
    \mathpalette{\da@xarrow{#1}{#2}{}\da@rightarrow{\,}{}}{}%
  }%
}
\newcommand{\xdashleftarrow}[2][]{%
  \mathrel{%
    \mathpalette{\da@xarrow{#1}{#2}\da@leftarrow{}{}{\,}}{}%
  }%
}
\newcommand*{\da@xarrow}[7]{%
  \sbox0{$\ifx#7\scriptstyle\scriptscriptstyle\else\scriptstyle\fi#5#1#6\m@th$}%
  \sbox2{$\ifx#7\scriptstyle\scriptscriptstyle\else\scriptstyle\fi#5#2#6\m@th$}%
  \sbox4{$#7\dabar@\m@th$}%
  \dimen@=\wd0 %
  \ifdim\wd2 >\dimen@
    \dimen@=\wd2 %
  \fi
  \count@=2 %
  \def\da@bars{\dabar@\dabar@}%
  \@whiledim\count@\wd4<\dimen@\do{%
    \advance\count@\@ne
    \expandafter\def\expandafter\da@bars\expandafter{%
      \da@bars
      \dabar@ 
    }%
  }%
  \mathrel{#3}%
  \mathrel{%
    \mathop{\da@bars}\limits
    \ifx\\#1\\%
    \else
      _{\copy0}%
    \fi
    \ifx\\#2\\%
    \else
      ^{\copy2}%
    \fi
  }%
  \mathrel{#4}%
}
\makeatother

\ifCLASSOPTIONcompsoc
  \usepackage[nocompress]{cite}
\else
  \usepackage{cite}
\fi

\ifCLASSINFOpdf
\else
\fi

\begin{document}
\title{UniFed: A Unified Framework for Federated Learning on Non-IID Image Features}

\author{Meirui~Jiang,
~\IEEEmembership{Student Member,~IEEE,}
Xiaoxiao~Li,
~\IEEEmembership{Member,~IEEE,}
Xiaofei~Zhang,
Michael~Kamp,
and~Qi~Dou,~\IEEEmembership{Member,~IEEE}

\IEEEcompsocitemizethanks{
\IEEEcompsocthanksitem M. Jiang and Q. Dou are with the Department
of Computer Science and Engineering, The Chinese University of Hong Kong.
\IEEEcompsocthanksitem X. Li is with the Department of Electrical and Computer Engineering, The University of British Columbia.
\IEEEcompsocthanksitem X. Zhang is with the School of Statistics and Mathematics, Zhongnan Univeristy of
Economics and Law.
\IEEEcompsocthanksitem M. Kamp is with the Institute for AI in Medicine (IKIM) at UK Essen,
Ruhr-University Bochum, and Monash University.
\IEEEcompsocthanksitem Corresponding author: Qi Dou (qidou@cuhk.edu.hk).}%

}

\markboth{Journal of \LaTeX\ Class Files,~Vol.~14, No.~8, August~2015}%
{Shell \MakeLowercase{\textit{et al.}}: Bare Demo of IEEEtran.cls for Computer Society Journals}

\IEEEtitleabstractindextext{%
\begin{abstract}
How to tackle non-iid data is a crucial topic in federated learning. This challenging problem not only affects training process, but also harms performance of clients not participating in training. Existing literature mainly focuses on either side, yet still lacks a unified solution to handle these two types (internal and external) of clients in a joint way. In this work, we propose a unified framework to tackle the non-iid issues for internal and external clients together. Firstly, we propose to use client-specific batch normalization in either internal or external clients to alleviate feature distribution shifts incurred by non-iid data. Then we present theoretical analysis to demonstrate the benefits of client-specific batch normalization. Specifically, we show that 
our approach promotes convergence speed for federated training and yields lower generalization error bound for external clients. Furthermore, we use causal reasoning to form a causal view to explain the advantages of our framework. At last, we conduct extensive experiments on natural and medical images to evaluate our method, where our method achieves state-of-the-art performance, faster convergence, and shows good compatibility. We also performed comprehensive analytical studies on a real-world medical dataset to demonstrate the effectiveness. 

\end{abstract}
}

\maketitle

\IEEEdisplaynontitleabstractindextext

\IEEEpeerreviewmaketitle

\IEEEraisesectionheading{\section{Introduction}\label{sec:introduction}}

\IEEEPARstart{F}{ederated} learning (FL) enables multiple institutions to collaboratively learn a joint global model without sharing data~\cite{ rieke2020future,wu2022communication,feddyn,sheller2020federated,dayan2021federated,dou2021federated,pati2022federated}. 
In FL, a common paradigm is federated averaging~\cite{fedavg}, in which clients train models locally and a server updates the global model by averaging parameters collected from the clients. While it has shown success in certain applications, recent studies~\cite{hsieh2020non, flchallenges,kairouz2021advances} reveal that federated methods perform less well on non-iid (not independent and identically distributed) data, significantly suffering from degradation of training convergence~\cite{li2019convergence}, performance~\cite{scaffold}, and generalization capability~\cite{liu2021feddg}. 
Therefore, designing new methods to tackle the data non-iid issue becomes imperative to unleash the potential of FL for real-world applications, in which the data are ordinarily not identically distributed, e.g., different imaging devices in healthcare, different environments (city, rural areas) for autonomous driving.

In FL systems, it is common that not all clients participate in the training process in each round. Some clients do not participate at all~\cite{kairouz2021advances,pati2022federated}. Current approaches to tackle non-iid data focus on those clients that either participate every round, or at least regularly during the training process -- we call those \textit{internal clients}. As we show in this paper, this, however, can deteriorate the performance on clients that do not participate in training -- the \textit{external clients}.
Recently, some federated generalization methods~\cite{liu2021feddg,fedadg,chen2022federated} propose to learn invariant features to improve the model generalization for external clients. 
Though with generalizability promoted, these methods are specifically designed for external clients, 
casting them to fit existing well-established FL algorithms would inevitably complicate the training process (e.g., sharing extra features~\cite{liu2021feddg,chen2022federated} or designing extra modules~\cite{fedadg}).
Therefore, exploring a unified framework considering both internal and external clients becomes essential.

Unlike most FL methods tackling non-iid data with label distribution shifts, we focus on the distribution shift in feature space. Non-iid image feature is a critical problem in real-world (medical imaging) and has not been fully studied in existing literature. 
This type of non-iid data typically happens in cases where the local devices are responsible for heterogeneity in the feature distributions. For example, in cancer diagnosis, the appearance of images collected from different hospitals varies dramatically due to different scanner vendors and imaging protocols~\cite{aubreville2020completely,liu2020ms,xu2021federated}, even though they have uniformly distributed labels (i.e., the cancer types treated are quite similar across the hospitals).

To tackle the non-iid data with feature shifts, feature normalization has been explored in centralized setting in the context of domain adaptation and generalization~\cite{li2016revisiting,huang2017arbitrary,chang2019domain,choi2021robustnet}. An effective approach is utilizing the Batch Normalization (BN)~\cite{ioffe2015batch}. Besides benefit in generalization, it has also demonstrated potential in improving optimization in centralized training~\cite{ioffe2015batch,santurkar2018does,benz2021revisiting}.
As an indispensable component of nearly all neural networks, BN can be readily composed in designing a unified framework, without complicating federated training or testing. Therefore, we aim to study the impact of BN layers in FL in order to assist in building a unified framework for non-iid features. However, either empirical study or theoretical analysis of BN in FL has not been fully studied yet.

Quite different from centralized paradigm, the unique property of distributed paradigm brings new challenges for analyzing models with BN layers in FL, including both training and testing stages. 
First, analysis for model training becomes difficult, because model update in each global round is influenced by all intermediate model states of internal clients. Specifically, the estimation of global population statistics in BN is performed in a distributed manner by aggregating locally estimated statistics. For each client, the data samples drawn from local distribution are not well described by the global statistics, which incurs normalized feature distribution deviating from the standard distribution. This deviation subsequently results in divergence between local and global model weights, making the analysis for training dynamics non-trivial. Second, when deploying model for testing, a global model trained on heterogeneous feature distributions fails to be an accurate estimation for each specific client, and it is even worse if the testing client does not present during training at all, i.e., external clients. For external clients, using either global statistics or internal clients' statistics for BN has a risk of improperly normalizing features, which deteriorates the final model prediction.How to measure the differences is crucial to understand and promote the expected performance for external clients.

The above-mentioned challenges are essentially coupled within BN, which should be handled jointly. In this paper, we present a unified framework (\textit{UniFed}) to solve these challenges together by proposing client-specific BN.  Client-specific BN is a lightweight yet effective method that each internal/external client uses its own BN layers. Our insights come from the inconsistency of statistic estimation in BN layers, i.e., the inconsistent local and global estimations during federated training, and the inconsistent training and testing estimations for external client testing.
By decomposing and analyzing the effects of BN in these two scenarios, we show that client-specific BN, as one single solution, can well promote FL internally and externally in terms of model performance, convergence and generalization capability.

For internal clients, using client-specific BN can well capture local data distribution statistics, alleviating the weight divergence induced by improperly normalized features. 
In our previous study~\cite{fedbn}, we investigated the training dynamics of overparameterized neural networks with client-specific BN through the neural tangent kernel~\cite{jacot2018neural}. In this work, we further improve the theoretical demonstrations to show that using client-specific BN improves the convergence rate. Then we extend our previous work to external clients to form a unified framework.

For external clients, due to the inconsistency between internal and external estimated statistics, using client-specific BN helps re-estimate statistics from external data. This re-estimation correlates external data distributions with its own statistics, thus promoting normalizing features into standard distribution. Such normalization well matches input space of other layers, effectively alleviating the model performance drop. 
By analyzing the generalization error on external data distribution, we theoretically demonstrate that client-specific BN decreases the divergence between distributions of internal and external clients, yielding a lower generalization bound than vanilla BN.

Furthermore, inspired by the resistance of human cognitive systems to feature distribution changes, we find humans immune to distribution changes because humans are more sensitive to causal relations than statistical associations~\cite{gopnik2004theory,geirhos2018generalisation}.
Therefore, we propose using causal reasoning~\cite{pearl2009causality,peters2017elements} to form a causal viewpoint to explain the benefits of client-specific BN. By constructing causal graphs for federated training and testing, we interpret client-specific BN as a specific distribution change resulting from the intervention on client's original feature distribution. This intervention helps identify causal relationships and omit nuisance factors (e.g., noises or different styles that usually make no impact on human decisions). The explanations integrate both federated training and testing stages, and well support our theoretical analysis on convergence rate and generalization error. 

Besides the theoretical analysis and explanations, we conduct extensive experiments to demonstrate the effectiveness of our proposed framework UniFed. The client-specific BN in this framework can be seen as a lightweight add-on for standard FL algorithms. First, we show the good compatibility of our method by applying client-specific BN on many different FL algorithms designed for non-iid data, including methods targeted at improving the global model from either client-side or server-side, and a method that aims to learn personalized models for each client. Then, we show that our method significantly boosts the performance for both internal and external clients. At last, we perform analytical studies to investigate the key properties of client-specific BN. Our experiments are performed on two datasets: one benchmark dataset of natural images with 15 different feature distribution shifts and one large-scale real-world medical dataset containing 450,000 histology patches.

It is noteworthy that some germane results of this work have been published in an earlier version in ICLR2021~\cite{fedbn}. However, our previous work only focuses on internal clients but lacks a unified view. This paper goes beyond the earlier version with significant improvements, including \textbf{(1)} proposing a unified framework to simultaneously improve federated training and testing on internal and external clients; \textbf{(2)} designing a novel client-specific BN method for external client testing; \textbf{(3)} theoretically demonstrating that client-specific BN yields a lower bound on generalization error compared to standard testing on external clients; \textbf{(4)} proposing a brand new causal viewpoint to explain our UniFed framework. \textbf{(5)} showing good compatibility of our method by applying it to eight different FL algorithms; \textbf{(6)} evaluating our unified framework on a large-scale real-world medical dataset and achieving state-of-the-art performance.

\section{Related Works}
This section focus on the FL methods on non-iid data, methods towards building a unified framework, and analysis of BN in neural networks in centralized training. These works form the motivation of our method.

\subsection{Federated Learning on Non-IID Data}
Recent FL studies have been increasingly investigating the issue of non-iid data distributions across different clients, from multiple aspects, such as federated training, client model personalization, and federated generalization. Most existing works aim to improve federated training, such as regularizing local model training~\cite{fedprox,feddyn,moon}, facilitating model optimization~\cite{scaffold,fedadam,feddc}, enhancing aggregation algorithm~\cite{fednova,pillutla2019robust}, improving feature normalization~\cite{fedbn,reisizadeh2020robust,silobn}, etc. 
For example, as an early work, FedProx~\cite{fedprox} extends the conventional FedAvg~\cite{fedavg} by adding a proximal term to help regularize the client training. SCAFFOLD~\cite{scaffold} suggests a new optimization algorithm to reduce variance for client drift in its local updates.
Later on, FedNova~\cite{fednova} proposes to use normalized stochastic gradients to perform global model aggregation rather than the cumulative raw local gradient changes; MOON~\cite{moon} proposes to use contrastive learning on latent feature representations to enhance the agreements between local and global models.
Very recently, FedDC~\cite{feddc} improved the SCAFFOLD~\cite{scaffold} by dynamically updating the objective function of each client, which contains a constraint penalty term and a gradient correction term.

Instead of learning a common global model, another promising direction for tackling non-iid data issue is learning personalized models. Personalized FL targets adapting the common consensus model to fit each client's individual data distribution. For example, PerFedAvg~\cite{perfedavg} introduces meta-learning in FL to obtain a personalized variant of FedAvg; Ditto~\cite{li2021ditto} utilizes regularization terms and multi-task learning objectives to train personalized models; FedBABU~\cite{fedbabu} proposes to share encoders across clients but preserve classifiers locally to perform the personalization.

Besides federated training on non-iid data, another important direction is federated generalization, which aims to improve the generalizability of federated models on unseen clients outside the federation. FedDG~\cite{liu2021feddg} first proposes this setting on medical images, and it presents a method to share client distribution information in a privacy-protecting way. FedADG~\cite{fedadg} extends federated generalization to natural images and proposes to use federated adversarial learning to measure and align distributions among different clients. Recently, Yuan et al.~\cite{yuan2022what} studied the generalization gap by considering two levels, i.e., out-of-sample gap and participation gap, and proposed a semantic synthesis strategy that enables realistic simulation without naturally-partitioned data; Caldarola et al.~\cite{fedsam} improve generalization in federated learning by employing sharpness-aware minimization and averaging stochastic weights.

Many efforts have been made to improve the FL on non-iid data, but all these existing works focus on boosting performance for clients either inside or outside the federation. There is still no unified solution to consider all these comprehensive scenarios jointly.

\subsection{Unified Framework for Multi-source Data}
Not only in the context of FL, the data heterogeneity issue has also been studied in many scenarios, such as multi-site learning, domain adaptation, and domain generalization. These works further motivate the emergency need for one unified framework to tackle the heterogeneity issue in internal and external data simultaneously. 
For example, in multi-site learning~\cite{liu2020ms,gauriau2021deep} works, experiments show that the model suffers a significant performance drop on external validation data. 
In domain adaptation and generalization which focus on improving performance for target (external) domain, a key idea is forcing a model to be domain agnostic. This idea essentially averages the domain conflict, thus the performance drops on each source (internal) domain~\cite{li2021dynamic}. Galstyan et al.~\cite{galstyan2022failure} also propose to measure metrics on training data when designing domain generalization algorithms.
These works only focus on internal training or external testing, but all emphasize the importance of designing well-performing algorithms on both internal and external data. Therefore, a unified framework is highly demanded.

\subsection{Batch Normalization Analysis in Neural Networks} 
Batch Normalization~\cite{ioffe2015batch} is an indispensable component in many deep neural networks and has shown its success in neural network training. Relevant literature has uncovered several benefits given by batch normalization. Santurkar et al.~\cite{santurkar2018does} show that BN makes the optimization landscape significantly smoother during training process, the smoothness induces a more predictive and stable behavior of the gradients. Luo et al.~\cite{luo2018towards} decompose BN into population normalization and gamma decay, and consider it an implicit regularizer to improve the robustness of optimization.
Yang et al.~\cite{yang2018a} provide a characterization of signal propagation and gradient backpropagation in wide BN networks at initialization.
Morcos et al.~\cite{morcos2018importance} suggest that BN implicitly discourages single direction
reliance, thus improving the model generalizability. Li et al.~\cite{li2018adaptive} take advantage of BN for tackling the domain adaptation problem. However, besides our early work~\cite{fedbn}, very limited study~\cite{silobn} considers the effects of BN in FL, and only empirical results are demonstrated. It is unclear what role BN plays in the scope of FL, especially for non-iid data. Comprehensive theoretical and empirical studies still remain unexplored to date.

\section{Method}
To address the non-iid features issue for both internal and external clients, we propose an effective new federated learning framework in a unified shape, named \textit{UniFed}. We start with the formulation of our method, then presents theoretical analysis to demonstrate the faster convergence rate, lower generalization error bound, and at last, use causal reasoning to explain our framework.

\subsection{Preliminaries}
We start with the formulation for FL on non-iid data with feature shifts. We assume $N\in\mathbb{N}$ clients jointly train for $T\in\mathbb{N}$ rounds. Denote the input space $\X \subset \R^d$ and the label space $\Y \subset \R$, where $d$ is the input dimension. Let $\b x \in \X$ and $y \in \Y$ be the input and output random variables following an underlying joint distribution $\D$ with density $p(\b x,y)$. 

For each client $i \in [N]$ ($[N] = \{1,\dots,N\}$), it has access to a training set of samples drawn from $\D_i$.
By rewriting $p_i(\b x,y)$ as $p_i(y|\b x)p_i(\b x)$ and $p_i(\b x|y)p_i(y)$, we define \textit{feature shift} as: 1) covariate shift: the $p_i(\b x)$ varies across clients, while $p_i(y|\b x)$ is the same for all clients; and 2) concept shift: the $p_i(\b x|y)$  varies across clients and $p(y)$ is the same. Let $f: \R^d \rightarrow \R$ be a model, each client aims to train its local model $f_i$ by minimizing the following risk:
\begin{equation}
\label{eq:client_risk}
{J}_{i}\left({f}_{i}\right):=\E_{(\b x, y) \sim p_i(\b x, y)}\left[\ell\left({f}_{i}(\b x), y\right)\right],
\end{equation}
where $\E$ denotes the expectation and $\ell$ is a proper loss function, e.g., the mean squared error loss function for the regression or the cross-entropy loss for the classification task. The overall goal of FL is that $N$ clients collaboratively train a global model $f$ by minimizing the aggregated risk:
\begin{equation}
\label{eq:fl_risk}
{J}\left({f}\right)=\frac{1}{N}\sum\nolimits_{i=1}^N J_i(f).
\end{equation}

\begin{figure*}[!t]
    \centering
    \includegraphics[width=0.9\textwidth]{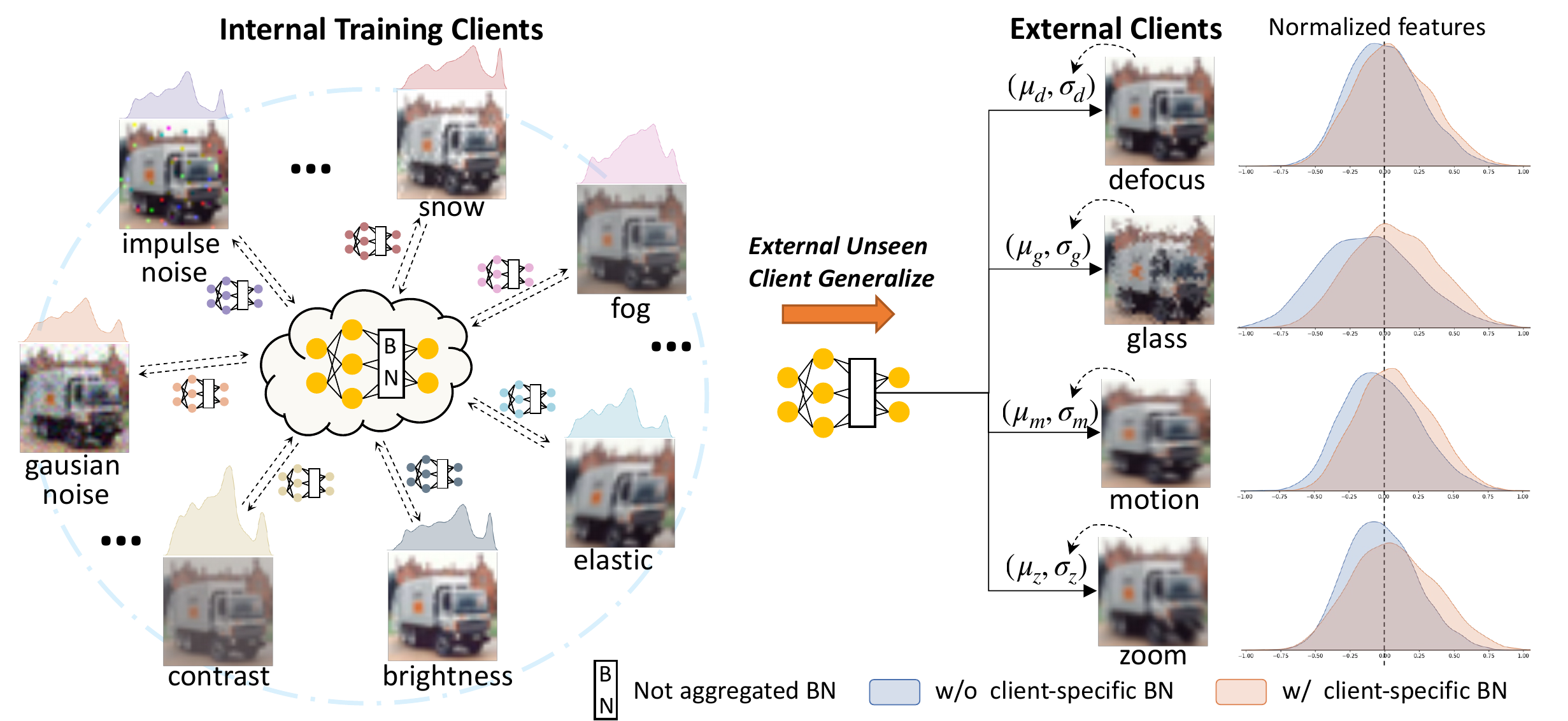}
    \caption{Overview of our proposed unified federated framework UniFed. Our framework considers two stages; for federated training on internal clients, client-specific BN means each client uses its local BN layers without joining the aggregation; for federated testing on external clients, client-specific BN are used by re-estimating the statistics from test data.}
    \label{fig:method}
\end{figure*}

\subsection{Proposed Unified Federated Framework -- UniFed}
The overall goal of FL is to train a global model $f$ that generalizes well with respect to $p(\b x,y)$. In practice, empirical risk minimization is commonly used as an approximation of $J_i(f_i)$ based on the client's local data. Since aggregated risk minimization is performed without sharing clients' local data, the non-iid features issue brings challenges from two aspects. First, the local risk minimization differs due to non-iid features, this diverges the aggregated risk minimization and harms the convergence rate. Second, the global model trained on heterogeneous features via the aggregated risk is not guaranteed to be an accurate estimation for each specific client, causing high generalization errors on testing clients. The error would largely increase when testing on external clients. It is a critical issue, because, in real-world settings, many clients may never participate in training~\cite{kairouz2021advances,yuan2022what}.

In this regard, we propose an efficient and effective framework UniFed to address the non-iid features issue. 
Our approach solves the joint risk minimization problem in Eq.(\ref{eq:fl_risk}) by tacking feature shifts via BN statistics calculated from each local client.
Fig.~\ref{fig:method} shows the overview of our framework, images over clients have non-iid features, and each client uses client-specific BN to tackle the non-iid features.
Our framework considers two stages: \textit{federated training} and \textit{federated testing}. We present the full algorithm in Algorithm~\ref{algo_box}. For federated training, each client preserves its own BN layers by excluding its parameters from the aggregation. For federated testing, if the client presented during training, we directly use the client model $w_i^T$, which only differs from the global model $w^T$ regarding client-specific BN parameters. If the client does not participate in training (a.k.a. external client), the client-specific BN strategy proposes re-estimating test data statistics for BN layers. Specifically, given an external client with batches of data to be tested sequentially (each batch has $n_t$ samples). The mean term $\mu$ and variance term $\sigma$ are calculated as:
\begin{equation}
\label{eq:mu_sigma}
\mu \gets \frac{1}{n_t}\sum\nolimits_{t=1}^{n_t} \b x_t,\quad
\sigma^2 \gets \frac{1}{n_t}\sum\nolimits_{i=1}^{n_t} (\b x_t-\mu)^2.
\end{equation}
Since the batch-level sampling may not accurately estimate the population-level mean and variance for a client, to stabilize the fluctuation, we further propose to use a momentum term $\tau$ to integrate relations among different batches:
\begin{equation}
\label{eq:momentum_mu_sigma}
\begin{gathered}
\E[\mu] \approx \mu \leftarrow \tau \mu + (1-\tau) \frac{1}{n_t}\sum\nolimits_{t=1}^{n_t} \b x_t,\\
\E[\sigma^2] \approx \sigma^2 \leftarrow \tau \sigma^2 + (1-\tau) \frac{1}{n_t}\sum\nolimits_{t=1}^{n_t} (\b x_t-\mu)^2.
\end{gathered}
\end{equation}
Our proposed client-specific BN results in significant empirical improvements in non-iid settings. We provide an explanation for these improvements by presenting theoretical analysis to show that our framework improves the convergence rate and yields a lower generalization error bound.

\begin{algorithm}[!t]
    \caption{\textbf{Our proposed unified federated framework}}
    \label{algo_box}
    \textbf{Notations:}
    {The client indexed by $i$, neural network layer indexed by $l$, initialized model parameters: $w_{i,l}^{0}$ for layer $l$ in client $i$, and total optimization round $T$.\\
    \textbf{Federated training:}\\
    \textbf{Output:} Local and global model parameters
    \begin{algorithmic}[1]
    \For{each round $t = 1, 2, \dots, T$ }
    \State{send the global model $w^t$ to each user k }
    \For{each client $i$ and each layer $l$ \textbf{in parallel}}
    \State{$w_{i,l}^{t+1} \gets SGD(w_{i,l}^{t})$} \Comment{Client training}
    \EndFor
    \For{each user $i$ and each layer $l$}
    \If{layer $l$ is not BatchNorm}
    \State{$w_{l}^{t+1} \gets \frac{1}{N}\sum_{i=1}^N w_{i,l}^{t+1}$} \Comment{Server aggregation}
    \EndIf
    \EndFor
    \EndFor\\
    \Return $\{w_{i}^T\}_{i=1}^N, w^T$
    \end{algorithmic} 
    \textbf{Federated testing:}\\
    \textbf{Output:} Prediction $\{\bar{y}_t\}_{t=1}^{n_t}$ for batch of test samples $\{\b x_t\}_{t=1}^{n_t}$
    \begin{algorithmic}[1]
    \If{Testing on \textit{internal clients}}
    \State{ $\{\bar{y}_t\}_{t=1}^{n_t}= f(\{\b x_t\}_{t=1}^{n_t};w_i^T)$}
    \EndIf
    \If{Testing on \textit{external clients}}
    \State{During the forward of $\{\bar{y}_t\}_{t=1}^{n_t}= f(\{\b x_t\}_{t=1}^{n_t};w^T)$}
    \For{each layer $l$ is BatchNorm}
    \State{Re-estimate test-specific statistics for BatchNorm}
    \State{$\mu \leftarrow \tau \mu + (1-\tau) \frac{1}{n_t}\sum\nolimits_{t=1}^{n_t} \b x_t$}
    \State{$\sigma^2 \leftarrow \tau \sigma^2 + (1-\tau) \frac{1}{n_t}\sum\nolimits_{t=1}^{n_t} (\b x_t-\mu)^2$}
    \EndFor
    \EndIf\\
    \Return $\{\bar{y}_t\}_{t=1}^{n_t}$
    \end{algorithmic} 
    }
\end{algorithm}

\subsection{Federated Training on Internal Clients with Faster Convergence}
Here we study the property of faster convergence by comparing our framework with and without client-specific BN. 
Our main idea is to study the training dynamics of networks through the neural tangent kernel (NTK)~\cite{jacot2018neural}. For simplicity, we assume all clients to have $M\in\mathbb{N}$ training examples for a regression task, i.e., each client $i \in [N]$ has training examples $\{(\b x^i_j, y^i_j)\in \R^d \times \R: j \in [M]\}$. 
Furthermore, we assume $f$ a two-layer ReLU neural network parameterized by $(\b W, \b c)\in \R^{m\times d}\times\R^m$ as:
\begin{equation}
\label{eq:fl_raw}
f(\b x; \b W, \b c) = 
\frac{1}{\sqrt{m}} \sum\nolimits^m_{k=1}c_k \phi(\b w_k^\top \b x),
\end{equation}
where $\phi(s) = \max\{s,0\}$ is the ReLU activation function. $\b w_k \in \R^d$ denotes the parameters of the first layer, where $k \in [m]$ and $m$ is the width of the hidden layer. $\b c$ is the top layer parameters of the network.
Let $\parallel\b w\parallel_{\b S} \triangleq \sqrt{\b w^\top \b S \b w}$ denote the induced vector norm for a positive definite matrix $\b S$. To investigate the gradient descent training with BN, we make the following assumption. 

\begin{assumption}[Data Distribution]
\label{as:data}
For each client $i \in [N]$ the inputs $\b x^i_j$ are centered $(\E \b x^i = \b 0)$ with covariance matrix $\b S_i = \E \b x^i \b x^{i\top} $, where $\b S_i$ is independent from the label $y$ and may differ for each $i\in [N]$ e.g., $\b S_i$ are not all identity matrices, and for each index pair $p \neq q$, $\b x_p \neq \kappa \cdot \b x_q$ for all $\kappa \in \R \setminus \{0\}$.
\end{assumption}

With Assumption \ref{as:data}, we incorporate the analysis on BN by drawing the following analogy, where batch training with BN amounts to:

\begin{equation}
\begin{aligned} 
\phi\!\left(\!\!\gamma\!\cdot\! \frac{\mathbf{w}\!^{\top}\!\mathbf{x}}{\sqrt{\mathbb{E}_{\mathbf{x}}\left(\mathbf{w}\!^{\top} \!\mathbf{x} \mathbf{x}\!^{\top} \!\mathbf{w}\right)}}\!\!\right)\!\! =\!\phi\!\left(\!\!\gamma\! \cdot \!\frac{\mathbf{w}\!^{\top} \!\mathbf{x}}{\sqrt{\mathbf{w}\!^{\top} \!\mathbf{S} \mathbf{w}}}\!\!\right) \!=\!\phi\!\left(\!\!\gamma\! \cdot \!\frac{\mathbf{w}\!^{\top} \!\mathbf{x}}{\|\mathbf{w}\| \mathbf{s}}\!\!\right)\!\!,
\end{aligned}
\end{equation}
where $\gamma$ is the scaling parameter of BN. 

Now we define our framework with client-specific BN parameters as $f^*: \R^d \rightarrow \R$ by re-parameterizing the Eq.(\ref{eq:fl_raw}) with $(\b V, \mathbf{\gamma}, \b c) \in \R^{m \times d} \times \R^{m \times N} \times \R ^m$, i.e., 
\begin{equation}
\begin{aligned} 
\label{eq:fedbn}
f^*(\b x; \b V, \mathbf{\gamma}, \b c) = 
\frac{1}{\sqrt{m}} \sum^m_{k=1}c_k \sum_{i=1}^N \phi\left(\gamma_{k,i} \cdot \frac{\b v_k^\top \b x}{\parallel \b v_k \parallel_{\b S_i}}\right) \cdot\\ \bm{1}\{\b x \in \text{client } i\} \enspace,
\end{aligned}
\end{equation}
where the normalization of the first layer for client $i$ is $ \frac{\b v_k^\top \b x^i}{\parallel \b v_k \parallel_{\b S_i}}$. Here, we omit learning the shift parameter of BN \footnote{
We omit centering neurons as well as learning the shift parameter of BN for the neural network analysis because of the assumption that $\b x$ is zero-mean and the two-layer network setting~\cite{kohler2019exponential,salimans2016weight}.}. If using aggregated BN, the framework instead trains a function $f: \R^d \rightarrow \R$ which is a special case of Eq.~(\ref{eq:fedbn}) with $\gamma_{k,i} = \gamma_{k}$ for $\forall i \in [N]$. We take a random initialization of the parameters \cite{salimans2016weight} in our analysis:
\begin{align}
\label{equ: inti}
\mathbf{v}_{k}(0) \sim N\left(0, \alpha^{2} \mathbf{I}\right), \quad c_{k} \sim U\{-1,1\}, 
\quad \\\text{and} \quad \gamma_{k} = \gamma_{k,i} = \left\|\mathbf{v}_{k}(0)\right\|_{2} / \alpha,
\end{align}
where $\alpha^2$ controls the magnitude of $\b v_k$ at initialization. The initialization of the BN parameters $\gamma_k$ and $\gamma_{k,i}$ are independent of $\alpha$. The parameters of the network $f^*(\b x; \b V, \mathbf{\gamma}, \b c)$ are obtained by minimizing the empirical risk derived from Eq.(\ref{eq:client_risk}). The loss function $\ell$ is chosen to be the squared loss, i.e., $\ell = (f^*(\b x_i)-y_i)^2$.

Then we study the trajectory of networks with aggregated BN ($f$) and with client-specific BN (ours) ($f^*$) through the NTK introduced by \cite{jacot2018neural}. Recent machine learning theory studies \cite{arora2019fine,du2018gradient,allen2019convergence,van2020training,dukler2020optimization} have shown that for finite-width over-parameterized networks, the convergence rate is controlled by the least eigenvalue of the induced kernel in the training evolution. 

We can decompose the NTK into a magnitude component $\mathbf{G}(t)$ and direction component $\mathbf{V}(t) / \alpha^{2}$ following \cite{dukler2020optimization}:
\begin{align*}
\frac{d \mathbf{f}}{d t}=-\mathbf{\Lambda}(t)(\mathbf{f}(t)- \b y), \quad \text{where} \quad
\mathbf{\Lambda}(t):=\frac{\mathbf{V}(t)}{\alpha^{2}}+\mathbf{G}(t).
\end{align*}
The specific forms of $\mathbf{V}(t)$ and $\mathbf{G}(t)$ are given in Appendix~\ref{sec:evol}. 
Let $\lambda_{min} (A)$ denote the minimal eigenvalue of matrix $A$. The matrices $\mathbf{V}(t)$ and $\mathbf{G}(t)$ are positive semi-definite, since they can be viewed as covariance matrices. This gives $
\lambda_{\min }(\mathbf{\Lambda}(t)) \geq \max \left\{\lambda_{\min }(\mathbf{V}(t)) / \alpha^{2}, \lambda_{\min }(\mathbf{G}(t))\right\}$.
According to NTK, the convergence rate is controlled by $\lambda_{min} (\mathbf{\Lambda}(t))$. 
Then, for $\alpha > 1$, convergence is dominated by $\b G(t)$. Let $\mathbf{\Lambda}(t)$ denote the evolution dynamics of network with aggregated BN, and $\mathbf{\Lambda}^*(t)$ denote our framework with client-specific BN. Let $\mathbf{G}(t)$ and $\mathbf{G}^*(t)$ denote the magnitude component in the evolution dynamics of using aggregated and client-specific BN.
For convergence analysis, we use the auxiliary version of Gram matrices, which is defined below. 
\begin{definition}
Given sample points $\{\b x_p\}_{p=1}^{NM}$, we define the auxiliary Gram matrices $\mathbf{G}^{\infty}  \in \R^{NM \times NM}$ and $\mathbf{G}^{*\infty}  \in \R^{NM \times NM}$ for framework with aggregated BN and client-specific BN as:
\begin{align}
\mathbf{G}_{p q}^{\infty} & := \mathbb{E}_{\mathbf{v} \sim N\left(0, \alpha^{2} \mathbf{I}\right)}
 \phi\left(\mathbf{v}^{\top} \mathbf{x}_{p}\right) \phi\left(\mathbf{v}^{\top} \mathbf{x}_{q}\right), \label{equ: fedavg_g} \\
 \mathbf{G}_{p q}^{*\infty} & := \mathbb{E}_{\mathbf{v} \sim N\left(0, \alpha^{2} \mathbf{I}\right)}
 \phi\left(\mathbf{v}^{\top} \mathbf{x}_{p}\right) \phi\left(\mathbf{v}^{\top} \mathbf{x}_{q}\right) \bm{1}\{i_p = i_q\}. \label{equ: fedbn_g}
\end{align}
\end{definition}
Given Assumption~\ref{as:data}, we use the key results in \cite{dukler2020optimization} to show that $\b G^{\infty}$ is positive definite. Further, we show that $\b G^{*\infty}$ is positive definite. We use the fact that the distance between $\b G(t)$ and its auxiliary version is small in over-parameterized neural networks, such that $\b G(t)$ remains positive definite.

\begin{lemma} 
\label{lemma:eigen}
Fix points $\{\b x_p\}_{p=1}^{NM}$ satisfying Assumption~\ref{as:data}. Then Gram matrices $\mathbf{G}^{\infty} $ and $\mathbf{G}^{*\infty}$ defined as in (\ref{equ: fedavg_g}) and (\ref{equ: fedbn_g}) are strictly positive definite. Let the least eigenvalues be $\lambda_{\min}(\b G^{\infty}) =: e_0$ and $\lambda_{\min}(\b G^{*\infty}) =: e^*_0,$ where $e_0,e_0^*>0$.
\end{lemma}

\textbf{Proof sketch}: The main idea follows \cite{du2018gradient,dukler2020optimization}, that given points $\{\b x_p\}_{p=1}^{NM}$, the matrices $\mathbf{G}^{\infty} $ and $\mathbf{G}^{*\infty}$ can be shown as covariance matrix of linearly independent operators. More details of the proof are given in the Appendix~\ref{sec:proof_lemma}.

Based on our formulation, the convergence rate of the framework using aggregated BN (Theorem~\ref{th:converge_g}) can be derived from \cite{dukler2020optimization} by considering non-identical covariance matrices. We derive the convergence rate of our framework in Corollary~\ref{th:converge_fedbn}. Our key result of comparing the convergence rates between using aggregated BN and client-specific BN is culminated in Corollary~\ref{coro:comp}.

\begin{theorem}[$\b G$-dominated convergence for aggregated BN~\cite{dukler2020optimization}]
\label{th:converge_g}
Suppose network (\ref{equ: fedavg_g}) is initialized as in (\ref{equ: inti}) with $\alpha >1$, trained using gradient descent and Assumptions \ref{as:data} holds. Given the loss function of training the neural network is the square loss with targets $\b y$ satisfying $\|\mathbf{y}\|_{\infty}=O(1)$. If $m=\Omega\left(\max \left\{N^{4}M^{4} \log (NM / \delta) / \alpha^{4} e_{0}^{4}, N^{2}M^2 \log (NM / \delta) / e_{0}^{2}\right\}\right)$, then with probability $1-\delta$,
\begin{enumerate}
\item For iterations $t = 0, 1, \cdots$, the evolution matrix $\mathbf{\Lambda}(t)$ satisfies $\lambda_{\min }(\mathbf{\Lambda}(t)) \geq \frac{e_{0}}{2}$.
\item Training with gradient descent of step-size $\eta = O\left(\frac{1}{\|\mathbf{\Lambda}(t)\|}\right)$ converges linearly as
\begin{align*}
\|\mathbf{f}(t)-\mathbf{y}\|_{2}^{2} \leq \left(1-\frac{\eta e_{0}}{2}\right)^{t}\|\mathbf{f}(0)-\mathbf{y}\|_{2}^{2}. 
\end{align*}
\end{enumerate}
\end{theorem}
Following the key ideas in \cite{dukler2020optimization}, here we further characterize the convergence for client-specific BN.  
\begin{corollary}[$\b G$-dominated convergence for client-specific BN]
\label{th:converge_fedbn}
Suppose network (\ref{equ: fedbn_g}) and all other conditions in Theorem \ref{th:converge_g}. With probability $1-\delta$, for iterations $t = 0, 1, \cdots$, the evolution matrix $\mathbf{\Lambda}^*(t)$ satisfies $\lambda_{\min }(\mathbf{\Lambda}^*(t)) \geq \frac{e^*_{0}}{2}$ and training with gradient descent of step-size $\eta = O\left(\frac{1}{\|\mathbf{\Lambda}^{*}(t)\|}\right)$ converges linearly as $\|\mathbf{f}^*(t)-\mathbf{y}\|_{2}^{2} \leq \left(1-\frac{\eta e^*_{0}}{2}\right)^{t}\|\mathbf{f}^*(0)-\mathbf{y}\|_{2}^{2}$.
\end{corollary}
The exponential factor of convergence for framework with aggregated BN $\left(1-\eta e_{0}/2\right)$ and client-specific BN $\left(1-\eta e^*_{0}/2\right)$ are controlled by the smallest eigenvalue of $
\b G(t)$ and $\b G^*(t)$,respectively. Then we can analyze the convergence performance by comparing $\lambda_{\min}(\b G^{\infty})$ and $\lambda_{\min}(\b G^{*\infty})$. 

\begin{corollary}[Convergence rate comparison between using aggregated BN and client-specific BN]
\label{co:46}
For the $\b G$-dominated convergence, the convergence rate of using client-specific BN is faster than that of using aggregated BN. 
\label{coro:comp}
\end{corollary} 
\textbf{Proof sketch}: The key is to show $\lambda_{\min}(\b G^{\infty}) \leq \lambda_{\min}(\b G^{*\infty})$. Comparing equation (\ref{equ: fedavg_g}) and (\ref{equ: fedbn_g}), $\b G^{*\infty}$ takes the $M \times M$ block matrices on the diagonal of $\b G^{\infty}$. Let $\b G^{\infty}_i$ be the $i$-th $M \times M$ block matrices on the diagonal of $\b G^{\infty}$. By linear algebra, $\lambda_{\min}(\b G^{\infty}_i) \geq \lambda_{\min}(\b G^{\infty})$ for $i \in [N]$. Since $\b G^{*\infty} = diag(\b G^{\infty}_1,\cdots,\b G^{\infty}_N)$, we have $\lambda_{\min}(\b G^{*\infty}) = \min_{i \in [N]} \{ \lambda_{\min}(\b G^{\infty}_i) \}$. Therefore, we have the result $\lambda_{\min}(\b G^{*\infty}) \geq \lambda_{\min}(\b G^{\infty})$.

\subsection{Federated Model Testing on External Clients with Lower Generalization Error}
\label{sec:bound}
In this section, we study the property of lower generalization error when applying our proposed unified framework for external clients. We start by considering the $\cH$-divergence~\cite{blitzer2007learning} between internal and external distributions, and the risk of $f$ on one client. Then we present the generalization error with $N$ training clients and derive the new lower error bound of UniFed. At last, we extend our analysis to multiple external clients. 
To study the generalization error, we consider two types of clients, \textit{seen} clients (i.e., the internal training clients) and \textit{unseen} clients (i.e., the external clients). The distribution for seen and unseen clients are denoted as $\D^s$ and $\D^u$, respectively. Recall that we have defined the risk of model $f$ for each client in Eq.~(\ref{eq:client_risk}), we denote the empirical risk as:
\begin{equation}
\label{eq:empirical_risk}
    \hat{J}_s(f):=\E_{(\b x,y)\sim\hat{\D}^s}[\ell(f(\b x), y)],
\end{equation}
where $\hat{\D^s}$ is the empirical distribution induced by samples drawn from $\D^s$. Similarly, the risk and empirical risk of $f$ on $\D^u$ and $\hat{\D^u}$ are denoted as $J_u(f)$ and $\hat{J}_u(f)$. Let $\cH$ be a hypothesis class for input space $\X$ such that $f\in\cH$, and $\A_{\cH}$ be the set of subsets of $\X$ that are the support of some hypothesis in $\cH$. Then the distance between two distributions is defined as: $d_{\cH}(\D,\D'):=2 \sup _{A\in \A_{\cH}}|P_{\D}(A) - P_{\D'}(A)|$.

The symmetric difference space $\cH \Delta \cH$ is defined as : $\cH \Delta \cH:=\{f(\b x) \oplus f'(\b x) | f,f' \in \cH\}$, where $\oplus$ is the XOR operation. $\A_{\cH \Delta \cH}$ is defined as the set of all sets $A$ such that $A = \{\b x : \b x \in \X, f(\b x) \neq f'(\b x)\}$ for some $f,f' \in \cH$ and it follows that a distance $d_{\cH \Delta \cH}$ is well-defined. We denote the \textit{ideal hypothesis} that minimizes combined seen clients and unseen clients risk as: $f^o = \argmin_{f \in \cH} J_s(f)+J_u(f)$ and the error of $f^o$ is defined as $\epsilon := J_s(f^o)+J_u(f^o)$.

We present the error bound of the unseen client given one seen client as following~\cite{albuquerque2019generalizing}: 
\begin{theorem}
\label{thm:initial_bound}
Let $\cH$ be a hypothesis space of VC-dimension $d$ and $\hat{\D^s}$, $\hat{\D^u}$ be samples of size $M$ each, drawn from $\D^s$ and $\D^u$, respectively. 
Let $\hat{d}_{\mathcal{H} \Delta \mathcal{H}}$ be the empirical distance induced by the symmetric difference hypothesis space.
With probability at least $1-\delta$ over the choice of the samples, for every $f \in \cH$,
\begin{equation}
\label{eq:initial_bound}
\begin{aligned}
    J_{u}(f) &\leq \hat{J}_{s}(f)+\frac{1}{2} \hat{d}_{\mathcal{H} \Delta \mathcal{H}}\left(\hat{\D^s}, \hat{\D^u}\right)
    \\&\quad+4 \sqrt{\frac{2 d \log \left(2M\right) +\log \left({4}/{\delta}\right)}{M}}+\epsilon.
\end{aligned}
\end{equation}
\end{theorem}

Considering the federated training with $N$ seen clients, the model we applied on unseen clients is actually the aggregation of models trained on seen clients. In this case, we assume $f_u = \sum_{i=1}^N\omega_i f_{s,i}$, $\forall \omega_i \in \R^{n}_{+}, \sum_{i=1}^n \omega_i = 1$. Then the distance $\hat{d}_{\mathcal{H} \Delta \mathcal{H}}$ can be further bounded as:
\begin{equation}
\hat{d}_{\mathcal{H} \Delta \mathcal{H}}(\hat{\D^s},\hat{\D^u}) \leq \sum\nolimits_{i=1}^N \omega_{i} \hat{d}_{\mathcal{H} \Delta \mathcal{H}}\left(\hat{\D^s_{i}}, \hat{\D^{u}}\right).
\end{equation}
The new upper bound incorporating federated clients is obtained by mainly applying the triangle inequality on the definition of distance $\hat{d}_{\mathcal{H} \Delta \mathcal{H}}$. For the detailed proofs please refer to Appendix~\ref{sec:proof_error_bound}. Similarly, we can transform the empirical risk $\hat{J}_s(f_u)$ and the error term $\epsilon$ as below:

\begin{equation}
    \hat{J}_s(f_u) = \hat{J}_s(\sum\nolimits_{i=1}^N \omega_i f_{s,i}), \quad \epsilon \leq \sum\nolimits_{i=1}^N \epsilon_i
\end{equation}

By replacing the terms  $\hat{d}_{\mathcal{H} \Delta \mathcal{H}}$, $\hat{J}_s(f_u)$ and $\epsilon$ into Eq.~(\ref{eq:initial_bound}), we further characterize the generalization bound with $N$ seen clients, which is presented in below.

\begin{corollary}[federated generalization bound for unseen client with aggregated BN]
\label{cor:error_bound_fl}
Under the assumptions of Theorem~\ref{thm:initial_bound}, let $\tilde{s}$ denote the mixture of source samples with size $NM$, and $\hat{d}_{\mathcal{H} \Delta \mathcal{H}}\left(\D_i^s, {\D}_{u}\right)$ denote the distance between the $i$-th seen client and the unseen client, the new generalization error trained with $N$ clients unseen is bounded as:
\begin{equation}
\label{eq:error_bound_avg}
\begin{aligned}
    J_{u}(f_{u}) &\leq \hat{J}_{\tilde{s}}(\sum\nolimits_{i=1}^N\!\omega_i f_{s,i})\!+\!\!\sum\nolimits_{i=1}^N \omega_i \!\!\left(\frac{1}{2} \hat{d}_{\mathcal{H} \Delta \mathcal{H}}\!\left(\hat{\D_i^s},\!\hat{\D^{u}}\right)\!+\!\epsilon_i\!\right)
    \\&\quad+4 \sqrt{\frac{2 d \log (2NM)+\log (4 / \delta)}{NM}}\enspace.
\end{aligned}
\end{equation}
\end{corollary}

This Corollary~\ref{cor:error_bound_fl} gives the error bound of an unseen client for the vanilla FL with aggregated normalization layers. Next, we consider our proposed framework that utilizes the client-specific BN strategy for the unseen client. Intuitively, the proposed client-specific BN performs the normalization using re-estimated mean and variance, making the unseen test data follow the standard distribution. Since the seen clients' distributions have been normalized to standard distribution, client-specific BN decreases the distance between distributions of seen and unseen clients. Then we formally present the new generalization error of our framework.

As we assume $\cH$ a hypothesis class of two-layer neural network with batch normalization, then the input is transformed as  $\z:=\frac{\b v_k^\top \b x^i}{\parallel \b v_k \parallel_{\b S_i}}$. With Assumption~\ref{as:data}, $\b x$ is Gaussian distributed, the transformed $\z$ follows $\N$, where $\N$ is the standard Gaussian distribution with mean $\mathbf{0}$ and covariance matrix as identity matrix.
For $\forall f \in \cH$, there exists a function $q$ such that $q(\z) = f(\b x;\b W)$ and for the true labeling function $g$, let $q_{t}(\z) = g(\b x;\b W_{t})$, where $\b W_{t}$ is the true model parameter. We have: 
\begin{equation}
\begin{aligned}
J_{s}(f) &= \E_{\b x \sim \N_i}[\ell(f(\b x),g(\b x))] \\&= \E_{\z \sim \N}[\ell(q(\z), q_t(\z))] = J_{\N}(q).
\end{aligned}
\end{equation}
For the seen clients, the risk becomes $J_{\N}(q)$ instead of $J_{s}(f)$, which means the covariate shift is eliminated by batch normalization. For unseen client with aggregated BN, the input is transformed as $\z_v:=\Sigma_s^{-1/2}(\b W \x - \mu_s)$, where $\mu_s$ and $\Sigma_s$ are the mean and variance matrix calculated using the seen clients data. While for unseen client with proposed client-specific BN strategy, the input is transformed as $\z_I:=\Sigma_t^{-1/2}(\b W \x - \mu_t)$, where $\mu_t$ and $\Sigma_t$ are the mean and variance matrix calculated using the unseen client data, i.e., the approximated estimation of true mean and variance for the unseen client in Eq.~(\ref{eq:momentum_mu_sigma}). Then we have $\z_I \sim \N$, this shows that the input samples drawn from training or testing distributions are transformed into the same standard Gaussian distribution $\N$ with the help of client-specific BN. The divergence term $\hat{d}_{\mathcal{H} \Delta \mathcal{H}}(\hat{\D^s},\hat{\D^u})$ between training and testing distributions can be further represented as $\hat{d}_{\mathcal{H} \Delta \mathcal{H}}(\N,\N)$, which can be eliminated. Following this, we present the new generalization error bound for UniFed, which uses the client-specific BN.

\begin{corollary}[federated generalization bound for unseen client with client-specific BN]
\label{cor:error_bound_fedbn}
Under the assumptions of Theorem~\ref{thm:initial_bound}, the new generalization error for UniFed with client-specific BN ($f^*$) is bounded as:
\begin{equation}
\label{eq:error_bound_bn}
\begin{aligned}
  J_{u}(f_u^*) & \leq 
  \hat{J}_{\tilde{\N}}(\sum\nolimits_{i=1}^N \omega_i f_i^*) + \sum\nolimits_{i=1}^N \omega_i \epsilon_i \\&\quad+4 \sqrt{\frac{2 d \log \left(2 NM\right)+\log \left({4}/{\delta}\right)}{NM}}.
\end{aligned}
\end{equation}
\end{corollary}

Comparing the generalization error by using aggregated BN (Eq.~(\ref{eq:error_bound_avg})) and ours (Eq.~(\ref{eq:error_bound_bn})), we can find our framework eliminates the empirical divergence term between training and testing distributions with the help of client-specific BN, yielding a lower error bound. Here we mainly show the results for one unseen client, and the results can be easily extended to multiple unseen clients. Considering $N_u$ unseen clients, the risk for unseen clients is bounded as:

\begin{equation}
    J_u(f^*_u) \leq \sum\nolimits_{i=1}^{N_u} J_{u_i} (f^*_u).
\end{equation}

By unrolling the risk terms for each unseen client, we can get the extended version of the generalization error bound for multiple unseen external clients. The detailed proofs for the theorem and corollaries are given in Appendix~\ref{app:error_bound}. 

\subsection{A Causal View of UniFed with Explainability}
Human cognitive systems are immune to the distribution change induced by different devices because humans are more sensitive to causal relations than statistical associations~\cite{gopnik2004theory}. The causal relation can be identified via causal reasoning~\cite{pearl2009causality,peters2017elements}, which introduces a core concept of intervention to ignore nuisance factors (i.e., not the cause of groundtruth).
Inspired by this, we propose to leverage causality to understand and explain the effects of UniFed.

For non-iid features, as the shifts of feature distributions (e.g., noises or different styles) usually make no impact on human decisions~\cite{dodge2017study,geirhos2018generalisation}, it is reasonable to assume that the feature shifts across clients come from nuisance factors. This is because the changes in task-related factors induce decision changes for both humans and neural networks.
For our proposed client-specific BN, it can be regarded as a specific distribution change resulting from the intervention on client's original feature distribution, which helps eliminate nuisance factors and preserves task-related causal features.

We first construct the causal graph, which is arguably the fundamental premise for causal reasoning~\cite{pearl2016causal,peters2017elements}.
Specifically, we construct two causal graphs to formalize the federated training and testing process using the following knowledge for analyzing feature normalization.
To study the effects of feature normalization, we propose to separate the features before and after normalization, and denote them as $R$ and $F$. Then we describe the feature extraction using neural network with normalization layers as  $X \rightarrow R \rightarrow F$, where raw feature $R$ is first extracted from input $X$, then normalization is applied to it to get normalized feature $F$. In general, feature extraction and normalization will be performed repeatedly, making a long chain start from $R$ and end at $F$. We instead use an arrow from $R$ to $F$ to simplify the graph without affecting the idea. In general, the finally calculated feature is relevant for the task we care about, i.e., $F \rightarrow Y$.
For each client, the training/inference happens at local side, where the input $X$ follows its local distribution $\D_i$ and the extracted feature is typically distribution-relevant~\cite{arjovsky2019invariant,piratla2020efficient,zhao2020domain}. This implies that $X \leftarrow \D \rightarrow R$. 
As federated learning distributedly train models for all clients and aggregate them to get a global model, it differs from training one model on centralized data. But when deploying model for testing, the data of a certain client is fully accessed.
Therefore, we construct two causal graphs $\mathcal{G}^{tr}$ and $\mathcal{G}^{te}$ for federated training and testing, by leveraging all this background knowledge. 

\begin{figure}[!t]
\centering
\includegraphics[width=\columnwidth]{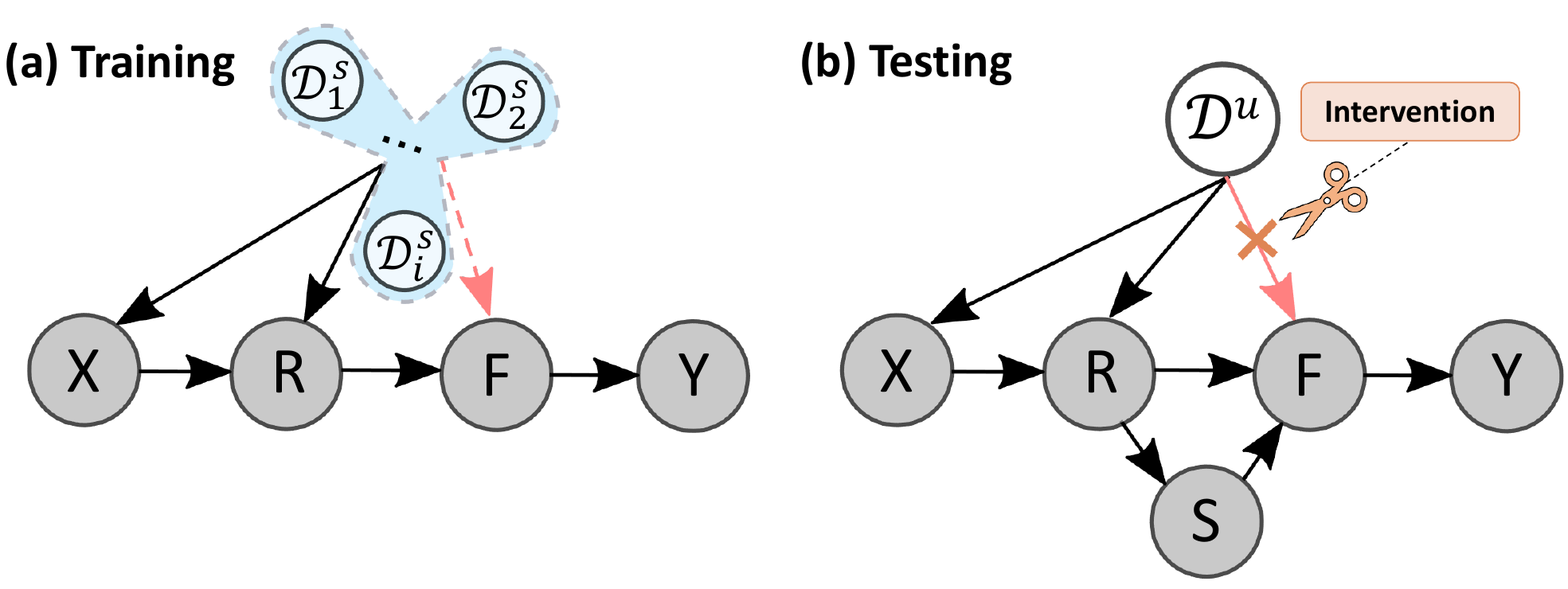}
\caption{The causal graphs for UniFed framework, including the training and testing two stages. Each node represents a random variable, and gray ones indicate observable variables. $X,R,F,Y,\D,S$ are input data, feature, normalized feature, label, distributions and test-specific statistics, respectively.}
\label{fig:causal}
\end{figure}

The causal graphs are shown in Fig.~\ref{fig:causal}, vertices denote variables and directed edges denote causal relations (e.g., $\emph{cause} \rightarrow \emph{effect}$). The gray ones indicate observable variables, i.e., the input data, features, and labels are observable during training/testing, while the underlying distributions for seen or unseen clients cannot be directly observed, we only access the samples drawn from distributions.
Note that besides the above-mentioned vertices and edges, there is an extra edge $\D \xdashrightarrow{} F$ and an extra node $S$. These elements are closely related to the properties of our UniFed framework, which will be detailed in later paragraphs.

\textbf{Causal graph for federated training.} 
For the causal graph of federated training, besides our previously mentioned structures, there is a dashed edge $\D \xdashrightarrow{} F$. This edge indicates the key differences between our client-specific BN and the standard method with aggregated BN. Specifically, with Assumption~\ref{as:data}, client-specific BN using locally estimated statistics helps each client normalize the features into standard Gaussian distribution. The normalized feature $F$ will no longer be statistically associated with its local underlying distributions. Here the client-specific BN is regarded as the intervention (denoted as $\textit{do}(F)$) resulting in normalized feature. While for aggregated BN, the essential step of statistics estimation in BN is performed in a distributed manner. The aggregated statistics do not accurately describe each client's local features. This results in features not being fully normalized into the standard distribution but still being relevant to local distributions. The differences in terms of $\D \xdashrightarrow{} F$ present evident effects in federated training. 

According to the causal graph $\mathcal{G}^{tr}$ depicted in Fig.~\ref{fig:causal}, for aggregated BN, there is a path, $X \leftarrow \D^s \rightarrow F \rightarrow Y$. This path shows that, given the improperly normalized features, labels are statistically correlated with the nuisance factors of distributions, which have no cause-effect on labels. This spurious correlation implies that the neural network is trained to fit the conditional association between labels and distributions. Therefore, the objective of local training with different distributions mismatches the aggregated objective function, deteriorating the model training.
While for our client-specific BN, without the presence of the edge from $D$ to $F$, the cause-effect of distribution variables only transit to $R$, but stopped at $F$ because those features are normalized to standard distribution. This property enables the model training in the UniFed framework not to be affected by heterogeneous features, showing faster convergence speed as we analyzed in the above section.

\textbf{Causal graph for federated testing.}
Here we further complete the causal graph for federated testing on unseen clients and analyze the effects of client-specific BN under this scenario. With the assumption of non-iid features, the unseen client holds a distribution $\D^u$ differing from training ones. Since BN layers from the global model are not trained, the features after normalization are inevitably associated with the unseen distribution $\D^u$. According to our previous discussion, the unseen distribution has no cause-effects on the groundtruth, the model performance would be affected if such nuisance factors are incorporated into features. 

We first discuss and analyze a typical solution in causal inference. And then analyze our proposed client-specific BN strategy, which is more simple and efficient compared with the typical solution. To remove nuisance factors (heterogeneous distributions in our work), causal inference proposes to use the back-door adjustment as a solution. Namely, let $P(Y|X)$ denote the classification model, and $P(Y|\emph{do}(X))$ denotes the new classifier that removes the various nuisance factors and pursues the true causal pattern determining the label of images. The back-door adjustment can be achieved by adjusting $X$ given different $D^s_i$ as below:

\begin{equation}
\label{eq:backdoor}
P(Y|\emph{do}(X))=\sum\nolimits_{i=1}^N P(Y|X, \D^s_i) P(\D^s_i).
\end{equation}
This back-door adjustment has been exploited in many works in the context of domain adaptation and domain generalization in centralized paradigm. Such as adjusting model to learn invariant features by generating adversarial samples from multiple source domains~\cite{zhao2018adversarial}, and using contrastive learning to adjust model learning invariant features across different domains~\cite{albuquerque2019generalizing}. However, casting these methods into FL is not trivial. A key of the adjustment in Eq.(\ref{eq:backdoor}) is that it needs to visit all the distributions (i.e., sum over the product of likelihood and prior of each distribution $\D^s_i$), which is not desired in FL with sensitive restrictions of data sharing. Some methods have tried to design federated generalization methods by sharing extra information like the image frequency~\cite{liu2021feddg} or specifically designed distribution generator~\cite{fedadg}.  But we argue that our client-specific BN strategy that addresses the unseen client generalization problem at test time is a more simple and less costly way than these methods. 

Similar to the client-specific BN in training, to remove the effects brought by the nuisance factor $\D^u$, a direct way is to use the unseen client-specific BN, i.e., using causal intervention on normalized features (i.e., $\emph{do}(F)$) to let the feature distribution similar to standard distributions. However, unlike the training process where the statistics have already been estimated, we do not have off-the-rack statistics estimations during testing. Therefore, we propose to accumulate statistics at the test-time with momentum, as shown in Eq.~(\ref{eq:momentum_mu_sigma}).
The statistics are calculated from the features, which can be formulated as introducing a surrogate variable $S$ into the causal graph, here the $S$ is test-specific statistics of raw features $R$ during testing.
With the help of the surrogate variable, we can achieve equivalent effects of intervening the normalized features s (i.e., \emph{$do(F)$}) by intervening the surrogate variable(i.e., \emph{$do(S)$}). Here the intervention on the surrogate variable is the operation of assigning the approximated estimation in BN layers.

We then present the causal language for interpreting show the equivalence of $do(S)$ and $do(F)$ in below Proposition~\ref{thm:intervention}.
For vertices $S, F, Y$ in our built graph $\mathcal{G}^{te}$, $S$ and $Y$ are independent given $F$ in $\mathcal{G}^{te}$, denoted as $(S \independent Y|F)_{\mathcal{G}^{te}}$.
Also, we denote $\mathcal{G}^{te}_{\overline{S}}$ ($\mathcal{G}^{te}_{\underline{S}}$) as the graph obtained by deleting from $\mathcal{G}^{te}$ all arrows pointing to (emerging from) nodes in $X$. 
\begin{proposition}
\label{thm:intervention} With the causal graph defined for testing, if a variable $S$ served as surrogate variable with following conditions satisfied:
(1)$F$ intercepts all directed paths from $S$ to $Y$;
(2)$P(Y|{\textit{do}}(S))$ is identifiable;
(3)$(Y \independent S|F)_{\mathcal{G}^{te}_{\overline{S} \overline{F}}}$; and 
(4)$(Y \independent F|S)_{\mathcal{G}^{te}_{\overline{S} \underline{F}}}$, 
then we have
\begin{equation}
\label{eq:internvention}
    P(Y|\textit{do}(F)) = P(Y|\textit{do}(S:=(\E[\mu_u],\E[\sigma_u^2]))).
\end{equation}
\end{proposition}

The proof of Proposition~\ref{thm:intervention} is shown in Appendix~\ref{app:proof_intervention}. 
The main idea is that the intervention on surrogate variable, i.e., $\textit{do}(S:=(\E[\mu_u],\E[\sigma_u^2]))$ (here $(\E[\mu_u],\E[\sigma_u^2])$ is assumed to be an unbiased estimation for the mean and variance pair of test-specific data), has the same effect on normalized feature $F$ as the passive observation $S=(\E[\mu_u],\E[\sigma_u^2])$ for every test sample from the unseen client.
By this analysis, we show the normalized features are intervened to follow the standard distribution. Such distribution transformation reduces the divergence between training and testing distribution, which is consistent with our generalization error analysis in the above section.

\begin{table*}[ht!]
\centering
\small
\setlength\tabcolsep{1.3pt}
\caption{Compatibility of our client-specific BN with existing FL algorithms. The cross sign indicates the improving point on which side compared with FedAvg.}\
\begin{tabular}{@{}ccccccccc@{}}
\toprule
& FedAvg~\cite{fedavg} & FedProx~\cite{fedprox} & Scaffold~\cite{scaffold} & FedAdam~\cite{fedadam} & FedNova~\cite{fednova} & MOON~\cite{moon} & FedDC~\cite{feddc} & PerFedAvg~\cite{perfedavg}  \\ \midrule
Client-side    &        & \plus & \plus & &         &  \plus &\plus & \plus  \\ \hline \\ [-1.7ex]
Server-side    &        &         &  & \plus & \plus &      &       &         \\[0.3ex] \hline \\ [-1.7ex]
\begin{tabular}[c]{@{}c@{}}Can be supported\\by our method?\end{tabular}&
\cmark       &    \cmark      &     \cmark      &     \cmark     &       \cmark   &    \cmark   &     \cmark   &       \cmark   \\ \bottomrule
\end{tabular}
\label{tab:fl_algos}
\end{table*}

\section{Experiments}
We validate our framework UniFed for internal federated client training and unseen external client testing on two datasets. We conduct extensive experiments to show that: 1) our method achieves a faster convergence rate than state-of-the-art FL algorithms during training, 2) our method improves the performance on internal clients, 3) our method improves the model generalizability on external clients, 4) how the external client testing results would be affected in real-world with ablation studies on three different key points. All the results are reported by calculating the average of three independent runs. Code will be released.

\begin{figure*}[ht]
    \centering
    \includegraphics[width=0.99\textwidth]{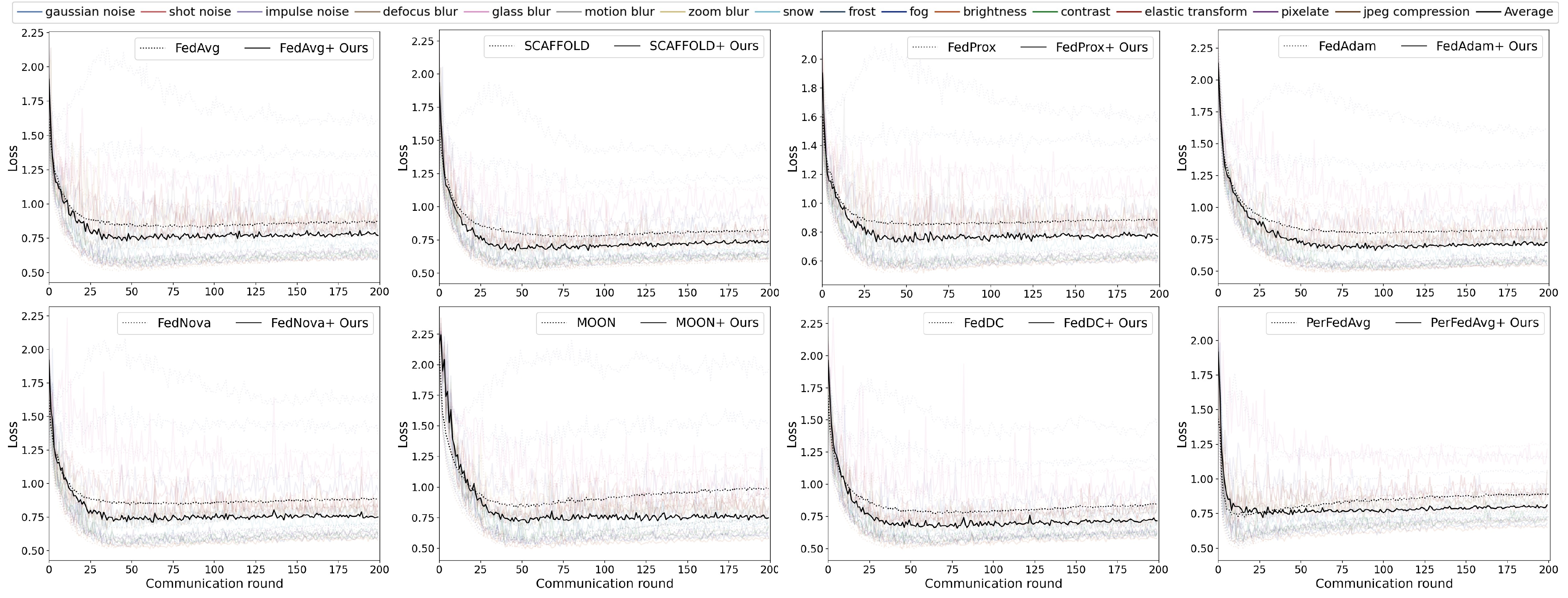}
    \vspace{-2.5mm}
    \caption{Convergence of FL algorithms with and without client-specific BN on the CIFAR-10-C dataset. Our method exhibits faster convergence and lower loss values.}
        
    \label{fig:convergence}
\end{figure*}

\begin{table*}[!ht]
\centering
\setlength{\tabcolsep}{0.5pt}
\small
\caption{Performance comparison of different FL methods before and after using our client-specific BN on the CIFAR-10-C dataset. Each column represents one internal client.}
\begin{tabular*}{\textwidth}{@{\extracolsep{\fill}\quad}lcccccccccccccccccc}
\hline\hline \\[-1.8ex]
Methods & A & B & C && D & E & F & G && H & I & J && K & L & M & N & O
\\ \midrule

\multirow{1}{*}{{FedAvg~\cite{fedavg}}} & 62.70  & 70.50 & 55.43 &\phantom{\tiny{,}}&
                80.49 & 64.25 & 74.83 & 71.15 &\phantom{\tiny{,}}& 78.32 & 78.38 & 80.42 &\phantom{\tiny{,}}& 81.05 & 79.75 & 73.78 & 79.55 & 76.47 \\

\multirow{1}{*}{+Ours \textbf{{\underline{+3.54}}}} & \textbf{76.72} & \textbf{74.83} & \textbf{71.72}         &\phantom{\tiny{,}}&
        \textbf{81.15} & \textbf{68.80} & \textbf{78.03} & \textbf{76.67} &\phantom{\tiny{,}}&
        \textbf{78.52} & \textbf{79.43} & \textbf{80.43} &\phantom{\tiny{,}}&
        \textbf{81.60} & \textbf{81.00} & \textbf{74.42} & \textbf{80.40} & \textbf{76.48} \\[0.1ex]
\hline \\[-1.9ex]

\multirow{1}{*}{{Scaffold~\cite{scaffold}}} & 66.15	& 73.12 & 58.52  &\phantom{\tiny{,}}
& 80.47 & 65.42 & 75.72 & 71.88 & \phantom{\tiny{,}}
& 79.02 & 79.43 & 80.83 &\phantom{\tiny{,}}
& {81.72} & 80.00 & 74.42 & 79.93 & \textbf{76.63} \\

\multirow{1}{*}{+Ours \textbf{{\underline{+3.19}}}} & \textbf{76.03} & \textbf{78.08} & \textbf{74.02} &\phantom{\tiny{,}}&
\textbf{81.45} & \textbf{71.45} & \textbf{77.95} & \textbf{77.92} &\phantom{\tiny{,}}&
\textbf{79.35} & \textbf{79.65} & \textbf{80.85} &\phantom{\tiny{,}}&
\textbf{81.78} & \textbf{80.55} & \textbf{75.47} & \textbf{80.00} & 76.48 \\
\hline \\[-1.9ex]

\multirow{1}{*}{{FedProx~\cite{fedprox}}} & 63.52 & 70.98 & 55.43 &\phantom{\tiny{,}}&
        80.48 & 63.88 & 74.78 & 70.68 &\phantom{\tiny{,}}&
        78.85 & 79.53 & 80.70 &\phantom{\tiny{,}}&
        \textbf{81.65} & 79.63 & 74.23 & 80.12 & \textbf{76.38} \\

\multirow{1}{*}{+Ours \textbf{{\underline{+3.41}}}} & \textbf{76.27} & \textbf{74.98} & \textbf{72.75} &\phantom{\tiny{,}}&
    \textbf{81.28} & \textbf{69.53} & \textbf{77.57} & \textbf{77.18} &\phantom{\tiny{,}}&
    \textbf{78.98} & \textbf{79.93} & \textbf{80.93} &\phantom{\tiny{,}}&
    81.62 & \textbf{80.48} & \textbf{74.42} & \textbf{80.13} & 75.98 \\ [0.1ex]
\hline \\[-1.9ex]

\multirow{1}{*}{{FedAdam~\cite{fedadam}}} & 62.87	& 71.57 & 55.82&\phantom{\tiny{,}}&
   81.32&	64.15&	75.55&	71.40&\phantom{\tiny{,}}&
   79.57&	79.67&	81.25&\phantom{\tiny{,}}&
    82.32&	80.53&	74.92&	80.52	&76.97 \\

\multirow{1}{*}{+Ours \textbf{{\underline{+3.55}}}} & \textbf{75.83} & \textbf{74.83} & \textbf{73.92} &\phantom{\tiny{,}}&
    \textbf{81.92} & \textbf{69.97} & \textbf{78.77} & \textbf{76.63} &\phantom{\tiny{,}}&
    \textbf{79.77} & \textbf{80.17} & \textbf{82.20} &\phantom{\tiny{,}}&
    \textbf{82.48} & \textbf{81.35} & \textbf{75.78} & \textbf{80.95} & \textbf{77.07} \\[0.1ex]
\hline \\[-1.9ex]

\multirow{1}{*}{{FedNova~\cite{fednova}}} & 62.45 &70.57&54.85&\phantom{\tiny{,}}&
    80.43&63.60&75.03&70.70&\phantom{\tiny{,}}&
    78.28&	78.52&	80.55&\phantom{\tiny{,}}&
    81.55&79.70&73.60&	79.80&	75.88\\

\multirow{1}{*}{+Ours \textbf{{\underline{+3.84}}}} & \textbf{76.30} & \textbf{75.53} & \textbf{73.03} &\phantom{\tiny{,}}&
    \textbf{81.25} & \textbf{69.18} & \textbf{78.00} & \textbf{76.05} &\phantom{\tiny{,}}&
    \textbf{78.58} & \textbf{80.02} & \textbf{81.30} &\phantom{\tiny{,}}&
    \textbf{81.90} & \textbf{80.97} & \textbf{74.22} & \textbf{80.37} & \textbf{76.37} \\[0.1ex]
\hline \\[-1.9ex]

\multirow{1}{*}{{MOON~\cite{moon}}} & 64.00	& 71.00 & 54.40 &\phantom{\tiny{,}}
& 81.33 & 64.07 &75.27 &71.63 &\phantom{\tiny{,}}
& 79.03 & 79.15 & 80.90 & \phantom{\tiny{,}}
& 81.88 & 80.00 & 73.88 &80.37 & 76.48\\

\multirow{1}{*}{+Ours \textbf{{\underline{+3.49}}}} & \textbf{75.92} & \textbf{75.73} & \textbf{72.78} &\phantom{\tiny{,}}&
\textbf{81.48} & \textbf{70.48} & \textbf{77.42} & \textbf{76.48} &\phantom{\tiny{,}}&
\textbf{79.07} & \textbf{79.78} & \textbf{81.55} &\phantom{\tiny{,}}&
\textbf{81.92} & \textbf{81.28} & \textbf{75.15} & \textbf{80.63} & \textbf{76.03}\\
\hline \\[-1.9ex]

\multirow{1}{*}{{FedDC~\cite{feddc}}} & 66.17 & 72.95 & 58.18 &\phantom{\tiny{,}}&
80.80 & 65.85 & 75.62 & 72.27 &\phantom{\tiny{,}}&
78.93 & 79.25 & 80.72 &\phantom{\tiny{,}}&
\textbf{81.82} & 79.73 & 74.12 & \textbf{80.55} & \textbf{76.83}\\

\multirow{1}{*}{+Ours \textbf{{\underline{+3.06}}}} & \textbf{76.63} & \textbf{77.62} & \textbf{74.27} &\phantom{\tiny{,}}&
\textbf{81.13} & \textbf{70.83} & \textbf{77.62} & \textbf{77.03} &\phantom{\tiny{,}}&
\textbf{79.32} & \textbf{79.95} & \textbf{81.37} &\phantom{\tiny{,}}&
81.65 & \textbf{80.65} & \textbf{75.20} & 80.53 & 75.93 \\  
\hline \\[-1.9ex]

\multirow{1}{*}{{PerFedAvg~\cite{perfedavg}}} & 76.83 & 80.10 & 76.33 &\phantom{\tiny{,}}&
83.50 & 73.53 & 79.72 & 78.12 &\phantom{\tiny{,}}&
81.43 & 80.77 & 82.92 &\phantom{\tiny{,}}&
83.63 & 82.52 & 78.00 & 82.67 & 79.85 \\

\multirow{1}{*}{+Ours \textbf{{\underline{+1.31}}}} & \textbf{80.75} & \textbf{80.87}  & \textbf{78.98} &\phantom{\tiny{,}}&
 \textbf{83.75} & \textbf{75.95} & \textbf{80.8}8 & \textbf{80.50} &\phantom{\tiny{,}}&
 \textbf{82.12} & \textbf{82.05} & \textbf{83.68} &\phantom{\tiny{,}}&
\textbf{83.82} & \textbf{83.60} & \textbf{79.63} & \textbf{83.00} & \textbf{79.93} \\   
\\[-1.8ex]\hline\hline 
\end{tabular*}
\label{table:cifar10c_fl_train}
\end{table*}

\subsection{Experiments on the Benchmark Dataset}
We have investigated many state-of-the-art FL algorithms designed for tackling non-iid data. No matter whether these methods are designed to overcome the non-iid issues from the client-side or server-side, we show that our client-specific BN strategy is compatible with these methods (shown in Table~\ref{tab:fl_algos}). This means our client-specific BN can be readily integrated into these methods to make further improvements.
We implement our proposed client-specific BN on current state-of-the-art FL algorithms to show the benefits of faster convergence, higher performance on internal clients, and better generalization performance on external clients. The methods include {FedAvg}~\cite{fedavg},
{Scaffold}~\cite{scaffold}
{FedProx}~\cite{li2018federated},  {FedAdam}~\cite{reddi2021adaptive},
{FedNova}~\cite{wang2020tackling},
{FedBN}~\cite{li2021fedbn},
{MOON}~\cite{moon},
{FedDC}~\cite{feddc} and 
{Per-FedAvg}~\cite{perfedavg}.

\subsubsection{Dataset and setup} We use the CIFAR-10-C dataset, which includes natural images with 15 different styles of corruptions which are grouped into four types (i.e., noise, blur, weather, digital)~\cite{hendrycks2019robustness}. Each corruption is regarded as a client. For the internal training, we use all 15 clients, and for external client testing, we use leave-one-CorruptionType-out validation (i.e., all corruption styles in three types form $D^s$ are used for training and validation, and all corruption styles of the remaining type are used for testing). This is designed to
keep a certain corruption type strictly unseen in training. The training client number is between 10 to 12. We use AlexNet~\cite{krizhevsky2012imagenet} pre-trained with the ImageNet and BN layers added after each convolution and fully-connected layer (as did in~\cite{hendrycks2019robustness}). We pre-process all the images by resizing them into $256\times256\times3$.

All our FL models are trained with cross-entropy loss and SGD optimizer. For data split, when performing experiments for internal federated training, all clients' data are randomly split into training, validation, and test sets with a ratio of 0.7, 0.1, and 0.2. As for the unseen external client testing, the data from internal clients are randomly split into 80\% for training and 20\% for validation for each client. The data from external clients are all held out for testing.
Without particular notice, we report the image classification accuracy results obtained with a batch size of 32 and a momentum value of 0.9. We train the model for 200 communication rounds with a learning rate of 0.001. The local update epoch of each client is set to 1. The model is selected based on the highest validation accuracy.

\begin{table*}[!th]
\centering
\setlength{\tabcolsep}{0.5pt}
\small
\caption{Performance of client-specific BN for generalization on CIFAR-10-C dataset. Each column represents one external client under leave-one-CorruptionType-out validation.}
\vspace{-2mm}
\begin{tabular*}{\textwidth}{@{\extracolsep{\fill}\quad}lcccccccccccccccccc}
\hline\hline \\[-1.8ex]
\multicolumn{1}{c}{\multirow{2}{*}{Methods}} & \multicolumn{3}{c}{\emph{noise}} & \phantom{\tiny{,}}& \multicolumn{4}{c}{\emph{blur}} &
\phantom{\tiny{,}} & \multicolumn{3}{c}{\emph{weather}} & \phantom{\tiny{,}} & \multicolumn{5}{c}{\emph{digital}} \\
\cmidrule{2-4} \cmidrule{6-9} \cmidrule{11-13} \cmidrule{15-19}
& N1 & N2 & N3 && B1 & B2 & B3 & B4 && W1 & W2 & W3 && D1 & D2 & D3 & D4 & D5 \\ \midrule

\multirow{1}{*}{{FedAvg~\cite{fedavg}}} & 58.30  & 71.09 & 30.46 &\phantom{\tiny{,}}&
                96.17 & 66.46 & 92.58 & 88.50 &\phantom{\tiny{,}}& 95.07 & 95.16 & 96.34 &\phantom{\tiny{,}}& \textbf{96.30} & 95.76 & 91.31 & 95.92 & 94.13 \\

\multirow{1}{*}{+Ours \textbf{{\underline{+9.62}}}} & \textbf{92.99} & \textbf{94.20} & \textbf{79.68}         &\phantom{\tiny{,}}&
        \textbf{96.29} & \textbf{87.33} & \textbf{95.79} & \textbf{95.70} &\phantom{\tiny{,}}&
        \textbf{95.77} & \textbf{96.02} & \textbf{96.38} &\phantom{\tiny{,}}&
        96.24 & \textbf{96.20} & \textbf{94.37} & \textbf{96.11} & \textbf{94.74} \\[0.1ex]
\hline \\[-1.9ex]
\multirow{1}{*}{{Scaffold~\cite{scaffold}}} & 62.05	& 74.52 & 36.24  &\phantom{\tiny{,}}
& 96.27& 70.64 &94.13 & 91.38 & \phantom{\tiny{,}}
& 95.20 & 95.42 & \textbf{96.33} &\phantom{\tiny{,}}
& \textbf{96.35} & {95.96} & 93.00 & {95.99} & \textbf{94.82} \\

\multirow{1}{*}{+Ours \textbf{{\underline{+8.30}}}} & \textbf{94.13} & \textbf{95.07} & \textbf{82.76} &\phantom{\tiny{,}}&
\textbf{96.30} & \textbf{88.27} & \textbf{95.67} & \textbf{95.52} &\phantom{\tiny{,}}&
\textbf{95.72} & \textbf{95.91} & {96.19} &\phantom{\tiny{,}}&
{96.18} & \textbf{96.10} & \textbf{94.35} & \textbf{95.99} & {94.77} \\
\hline \\[-1.9ex]

\multirow{1}{*}{{FedProx~\cite{fedprox}}} & 58.42 & 71.21 & 31.68 &\phantom{\tiny{,}}&
        96.22 & 64.67 & 91.83 & 87.28 &\phantom{\tiny{,}}&
        95.00 & 95.04 & 96.34 &\phantom{\tiny{,}}&
        \textbf{96.35} & 95.70 & 90.69 & 95.98 & 93.88 \\

\multirow{1}{*}{+Ours \textbf{{\underline{+9.72}}}} & \textbf{92.65} & \textbf{94.15} & \textbf{79.30} &\phantom{\tiny{,}}&
    \textbf{96.34} & \textbf{86.37} & \textbf{95.70} & \textbf{95.65} &\phantom{\tiny{,}}&
    \textbf{95.75} & \textbf{95.99} & \textbf{96.42} &\phantom{\tiny{,}}&
    96.28 & \textbf{96.26} & \textbf{94.36} & \textbf{96.16} & \textbf{94.74} \\ [0.1ex]
\hline \\[-1.9ex]

\multirow{1}{*}{{FedAdam~\cite{fedadam}}} & 58.25	&70.78&	32.41&\phantom{\tiny{,}}&
    96.37&	67.97&	92.81&	88.85&\phantom{\tiny{,}}&
    93.70&	93.62&	95.63&\phantom{\tiny{,}}&
    96.41&	95.87&	90.87&	96.05	&93.92 \\

\multirow{1}{*}{+Ours \textbf{{\underline{+9.71}}}} & \textbf{93.24} & \textbf{94.53} & \textbf{80.40} &\phantom{\tiny{,}}&
    \textbf{96.48} & \textbf{87.58} & \textbf{95.89} & \textbf{95.85} &\phantom{\tiny{,}}&
    \textbf{95.11} & \textbf{95.60} & \textbf{96.15} &\phantom{\tiny{,}}&
    \textbf{96.45} & \textbf{96.37} & \textbf{94.38} & \textbf{96.23} & \textbf{94.87} \\[0.1ex]
\hline \\[-1.9ex]

\multirow{1}{*}{{FedNova~\cite{fednova}}} & 75.82&	84.79&	32.62&\phantom{\tiny{,}}&
    96.69&	75.69&	94.85&	92.88&\phantom{\tiny{,}}&
    95.78&	95.47&	96.66&\phantom{\tiny{,}}&
    \textbf{96.83}&	96.42&	94.43&	96.59&	95.75\\

\multirow{1}{*}{+Ours \textbf{{\underline{+7.21}}}} & \textbf{95.55} & \textbf{96.09} & \textbf{86.10} &\phantom{\tiny{,}}&
    \textbf{96.75} & \textbf{91.37} & \textbf{96.24} & \textbf{96.20} &\phantom{\tiny{,}}&
    \textbf{96.39} & \textbf{96.49} & \textbf{96.76} &\phantom{\tiny{,}}&
    {96.81} & \textbf{96.71} & \textbf{95.53} & \textbf{96.64} & \textbf{95.82} \\[0.1ex]
\hline \\[-1.9ex]

\multirow{1}{*}{{FedBN~\cite{fedbn}}} & 55.87	&68.84	&28.44	&\phantom{\tiny{,}}&
    96.04	&64.56	&90.38	&84.95	&\phantom{\tiny{,}}&
    94.38	&94.42 & 96.06	&\phantom{\tiny{,}}&
    96.11&	94.61&	87.26&	95.47&	91.99\\

\multirow{1}{*}{+Ours \textbf{{\underline{+10.60}}}} & \textbf{91.53} & \textbf{93.01} & \textbf{77.30} &\phantom{\tiny{,}}&
    \textbf{96.42} & \textbf{85.48} & \textbf{95.71} & \textbf{95.52} &\phantom{\tiny{,}}&
    \textbf{95.46} & \textbf{95.87} & \textbf{96.35} &\phantom{\tiny{,}}&
    \textbf{96.30} & \textbf{96.17} & \textbf{93.48} & \textbf{96.06} & \textbf{93.77} \\
    \hline \\[-1.9ex]
    
\multirow{1}{*}{{MOON~\cite{moon}}} & 62.45	& 75.05 & 35.14 &\phantom{\tiny{,}}
& 96.33 & 66.80 & 93.67 & 90.71 &\phantom{\tiny{,}}
& 95.40 & 95.38 & 96.31 & \phantom{\tiny{,}}
& {96.27} & 95.76 & 92.30 & 95.97 & 94.65\\

\multirow{1}{*}{+Ours \textbf{{\underline{+8.79}}}} & \textbf{94.12} & \textbf{95.07} & \textbf{82.58} &\phantom{\tiny{,}}&
\textbf{96.44} & \textbf{87.85} & \textbf{95.82} & \textbf{95.74} &\phantom{\tiny{,}}&
\textbf{95.96} & \textbf{96.02} & \textbf{96.35} &\phantom{\tiny{,}}&
\textbf{96.27} & \textbf{96.19} & \textbf{94.50} & \textbf{96.12} & \textbf{95.00} \\
\hline \\[-1.9ex]

\multirow{1}{*}{{FedDC~\cite{feddc}}} & 62.63 & 74.53 & 33.86 &\phantom{\tiny{,}}&
96.27 & 70.72 & 94.18 & 91.58 &\phantom{\tiny{,}}&
95.21 & 95.49 & {96.31} &\phantom{\tiny{,}}&
{96.31} & 95.89 & 93.04 & 96.04 & 94.85 \\

\multirow{1}{*}{+Ours \textbf{{\underline{+8.54}}}} & \textbf{93.95} & \textbf{95.05} & \textbf{82.73} &\phantom{\tiny{,}}&
\textbf{96.35} & \textbf{88.72} & \textbf{95.89} & \textbf{95.65} &\phantom{\tiny{,}}&
\textbf{95.84} & \textbf{96.03} & \textbf{96.32} &\phantom{\tiny{,}}&
\textbf{96.31} & \textbf{96.22} & \textbf{94.80} & \textbf{96.04} & \textbf{95.09} \\  
\hline \\[-1.9ex]

\multirow{1}{*}{{PerFedAvg~\cite{perfedavg}}} & 85.82 & 93.69 & 54.99 &\phantom{\tiny{,}}&
96.59 &82.36 &95.43 &94.91 &\phantom{\tiny{,}}&
 96.07&96.17 &96.68 &\phantom{\tiny{,}}&
96.81 & 96.47 &95.14 &96.59 & 95.77 \\

\multirow{1}{*}{+Ours \textbf{{\underline{+3.90}}}} & \textbf{95.76} & \textbf{96.17} & \textbf{87.70} &\phantom{\tiny{,}}&
 \textbf{96.74}& \textbf{92.07}& \textbf{96.26} & \textbf{96.14} &\phantom{\tiny{,}}&
 \textbf{96.35}& \textbf{96.47} & \textbf{96.74} &\phantom{\tiny{,}}&
\textbf{96.81} & \textbf{96.67} &\textbf{95.64} & \textbf{96.60}&\textbf{95.81}  \\   
\\[-1.8ex]\hline\hline 
\end{tabular*}
\label{table:cifar10c_fl_test}
\end{table*}

\begin{figure}[!t]
    \centering
    \includegraphics[width=0.99\columnwidth]{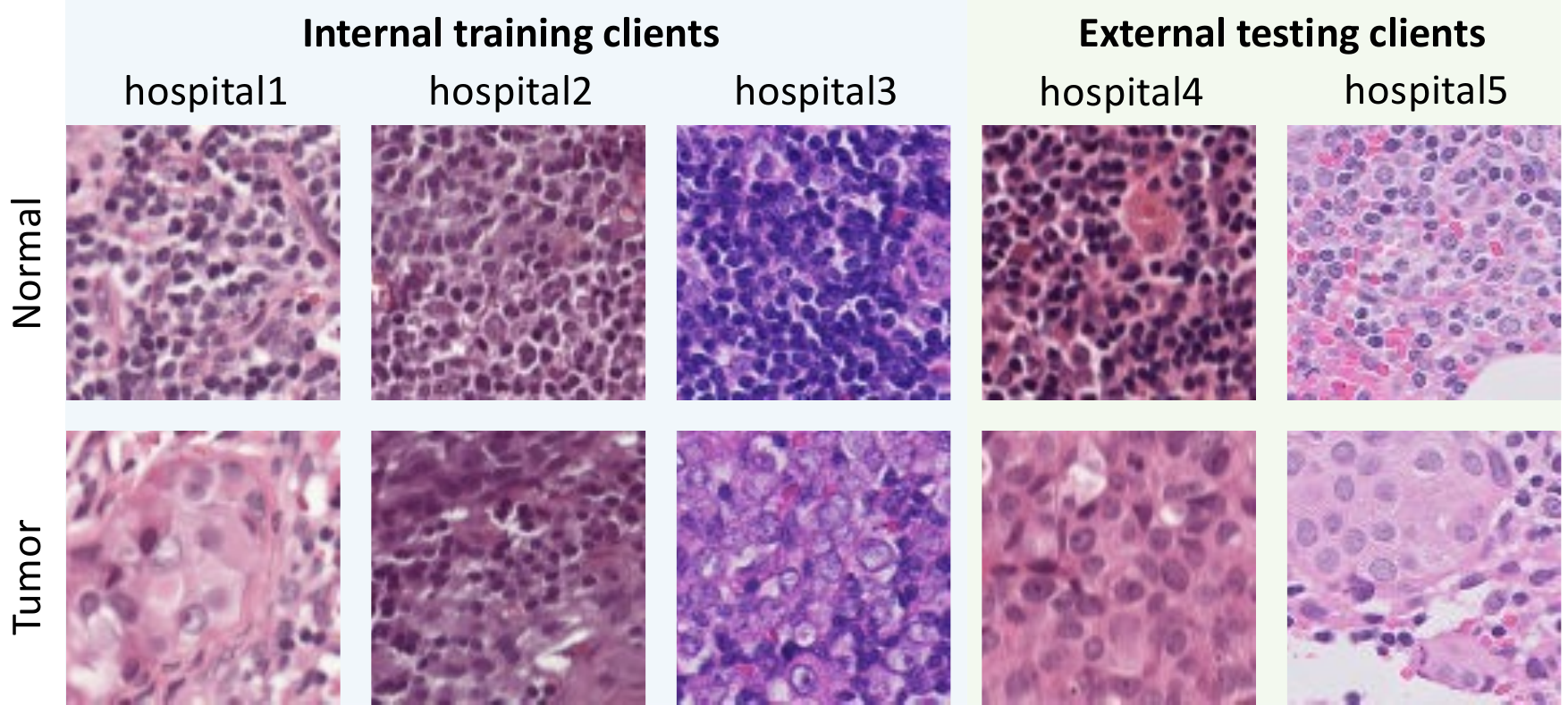}
        \caption{Examples of breast histology images of normal and tumor tissues from five clients, showing large heterogeneity.}
    \label{fig:camelyon_sample}
\end{figure}

\subsubsection{Internal federated training}
For each FL algorithm, we perform the comparison by using: 1) original algorithms for federated training, 2) our proposed client-specific BN for federated training.

We first present the convergence speed during training. The results are shown in Fig.~\ref{fig:convergence}. We plot the loss curves for each client and the average loss. The original algorithm is denoted in a dashed line, and client-specific BN (+Ours) is denoted in a solid line. From the figure can be observed that, our proposed strategy clearly improves the average convergence speed for almost all of the FL algorithms, i.e., achieves the same loss with fewer communication rounds. Besides, client-specific BN achieves lower loss than original methods when the model has converged. This attributes to the non-iid features are normalized into a standard distribution by client-specific BN, alleviating the divergence for the global model during model aggregation. Furthermore, we notice that some methods (e.g., MOON, PerFedAvg) may present the overfitting issue (i.e., the loss curve increases at late training rounds), while our strategy keeps a stable training trend. This observation also demonstrates the benefits of using client-specific BN for federated training.
\\\indent We then show the performance improvements for internal clients in Table~\ref{table:cifar10c_fl_train}. In this table, the first row of each group is the testing performance using original algorithms, and the second row is the performance using our client-specific BN. Some methods show better performance compared the baseline method FedAvg. On top of this, our approach consistently improves by at least 3\% on the average test accuracy for almost all methods. Note that PerFedAvg is a personalized method, this approach further updates model by performing extra SGD update steps using several labeled test data. Our client-specific BN consistently improves PerFedAvg on every single client, even though the performance boost is not as large as others. By including eight state-of-the-art FL methods for comparison, the overall efficacy and good compatibility of client-specific BN are well validated.  

\subsubsection{External client testing}

\begin{figure*}[!t]
    \centering
    \includegraphics[width=\textwidth]{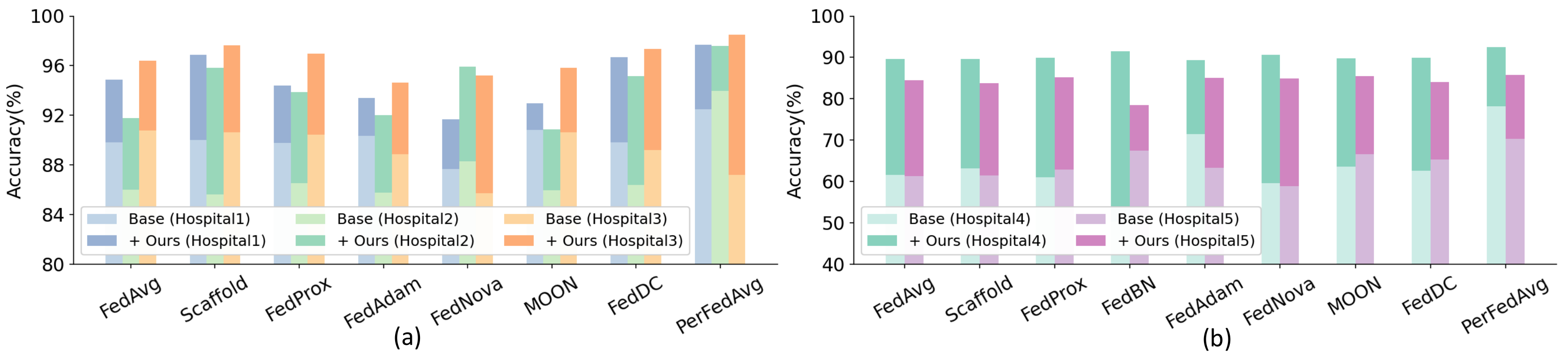}
    \caption{(a) and (b) show the results on the medical image dataset with three internal training clients and two external testing clients. Base refers to the original algorithm, and ours refer to the client-specific BN.}
    \label{fig:camelyon_result}
\end{figure*}

\begin{figure*}[!t]
    \centering
    \includegraphics[width=1.8\columnwidth]{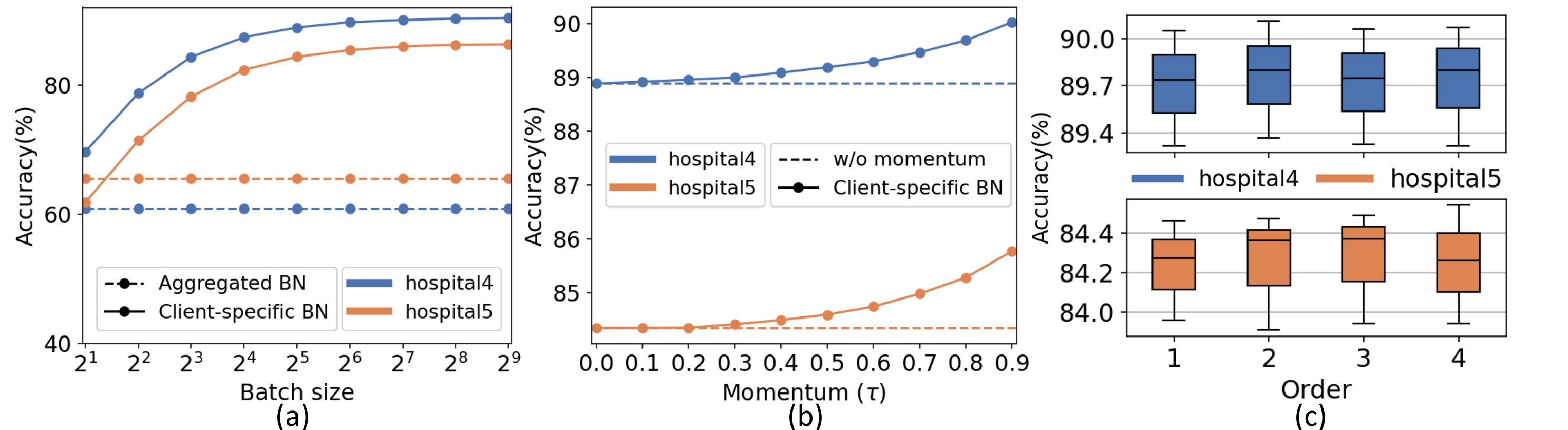}
    \caption{(a) and (b) study how the batch size and momentum values would affect client-specific BN. (c) is the performance comparison of client-specific BN using different sampling sequences.}
    \label{fig:abliation}
\end{figure*}

Besides the validation of federated training, we further present the benefits for improving performance on unseen external clients by using client-specific BN. Specifically, we compare two methods: 1) directly deploying the global model trained by different FL methods for testing, 2) using our proposed client-specific BN by re-calculating test-specific statistics with a momentum term. Table~\ref{table:cifar10c_fl_test} show the results, and the FedBN denotes the variant of FedAvg by using client-specific BN during training, i.e., the BN layers in the global model are not updated. 
It is noticed that, our approach achieves a significant increase in unseen client testing accuracy on the challenging CIFAR-10-C dataset, which presents a wide scope of feature shift. On average, for all unseen testing clients, our obtained unseen client generalization results increase at least 7.21\% for non-personalized FL methods, and further increase accuracy by 3.9\% on the PerFedAvg, which utilizes a small number of labeled test data. 

Overall speaking, the results on the benchmark dataset, which presents 15 different types of feature shifts, well demonstrate the benefits (i.e., faster convergence, higher performance, and better generalizability) of our method on internal and external clients in both federated training and testing scenarios. 

\subsection{Experiments on Real-world Medical Image Dataset}
To better understand how our framework can be beneficial in real-world non-iid features problems, we have extensively validated UniFed on a large-scale real-world medical dataset.

\subsubsection{Dataset and setup}
We use the public tumor classification dataset Camelyon17, which shows histology images with different stains from 5 hospitals~\cite{bandi2018detection}. The samples are shown in Fig.~\ref{fig:camelyon_sample}.
Following the setting of source/target domains in literature~\cite{bandi2018detection,koh2020wilds}, we use three hospitals (named hospital1,2 and 3) for internal federated training, and test on two hospitals as external clients (denoted as hospital4 and hospital5). We use DenseNet121~\cite{huang2017densely} and keep other training settings the same as the benchmark experiment. This medical dataset validates our method on a large-scale (i.e., 4.5 million images) real-world healthcare scenario, demonstrating its potential for practical applications.

\subsubsection{Results and Analysis}
The experiment results for federated training and testing are shown in Fig.~\ref{fig:camelyon_result}. For internal clients, the performance after applying client-specific BN is significantly improved by at least 4\%, and for external clients, the accuracies are surprisingly boosted by at least 14\% on average. Interestingly, we find the performance gap between personalized and non-personalized FL methods is mitigated by applying our client-specific BN strategy. The results are inspiring and promising for deploying UniFed to the healthcare field, where data are often limited, and feature distributions are heterogeneous.

\subsubsection{Analytical studies}
Our client-specific BN has no extra hyperparameters to tune during training, while for testing, there might be some factors related to the prediction quality. We conduct ablation studies on the real-world medical dataset to investigate three points, including batch size, momentum value, and the fluctuation of statistics with different sampling orders.

\textbf{One-batch experiment with different batch sizes.} We conduct the one-batch study, which explores how batch size can affect model performance on external clients with only one sample batch (i.e., no mean or variance tracked with momentum). As shown in Fig.~\ref{fig:abliation}(a), we use aggregated BN as the default baseline and test the model on entire data of external clients with different batch sizes (from 2 to 512). The baseline (dotted line) using fixed training-estimated BN is not influenced by batch size, but for our proposed client-specific BN, as expected, the test accuracy increases as we enlarge the batch size, because more samples assist the representative estimation of statistical distributions. Meanwhile, we see that the accuracy becomes saturated near the highest performance after the batch size is sufficiently large (e.g., reaching 32).

\textbf{Effects of momentum value.}
We further study whether the FL model performance is sensitive to the momentum value of $\tau$, as results are shown in Fig.~\ref{fig:abliation}(b). In this experiment, we set the test batch size as 32, and change the value of momentum between $[0.0,0.9]$ (noting that $\tau=0$ means no historical values are tracked). It is observed that, within a wide range, using momentum is beneficial to performance improvement, especially when $\tau>0.5$, while the overall performance is not very sensitive to the specific value of momentum (with accuracy changing within around 1\%).

\textbf{Sensitivity of sampling order.}
To study if the sampling order would affect the estimation during test time, we use the box plot to demonstrate the test accuracy with different sampling orders. For each order, we report the results of models trained with different random seeds. From Fig.~\ref{fig:abliation}(c) can be observed that our proposed client-specific BN has only slight changes under different orders, showing the potential under real applications that have no specific sampling sequence.

\section{Discussion}
The federated model may suffer different degrees of performance degeneration with the presence of heterogeneity in feature space.
This problem becomes even more severe for external clients with unseen data distribution shift. It is imperative to overcome the heterogeneity to promote federated model quality, especially for critical medical applications with a low error tolerance~\cite{liu2022medical}.
Our proposed unified federated framework can mitigate the heterogeneity in feature space and help improve the training convergence as well as performance on both internal and external clients. Although many efforts have been put into designing FL algorithms for non-iid data, our framework first bridges the gap between internal and external clients with one light-weight and effective strategy. Besides showing effectiveness from empirical results, we also provide theoretical analysis for internal training and external testing stages, and explain the whole framework from a view of causal reasoning.

\textbf{Compatibility and complexity.}
Our framework is designed in a unified shape by one client-specific BN strategy, which can be seen as a lightweight add-on to be readily combined with existing FL methods. This strategy is independent of the optimization, communication, and aggregation designs in FL and only requires the model to contain the BN layer, which is an indispensable component of nearly all neural networks. Our experimental results have demonstrated the good compatibility of client-specific BN on various FL algorithms which have different method designs. The theoretical analysis of such a combination is an interesting direction for future work. We also note that since our framework does not require BN layers to be included in aggregation, this helps reduce the communication burden.

\textbf{Privacy concerns.}
Although the FL paradigm reduces the privacy risks by keeping data locally, recent works demonstrated that the sharing of model parameter updates might also make FL vulnerable to gradient inversion attacks~\cite{geiping2020inverting,hatamizadeh2022gradient}. However, due to the property of not aggregating BN layers, our framework presents extra merit for reducing the risks of such gradient inversion attacks in FL~\cite{huang2021evaluating}. Specifically, since the exposure of statistics in BN layers significantly improves the quality of gradient inversion, our client-specific BN strategy weakens the quality of attacks by not sharing the statistics. And our strategy also does not affect the final model performance.

\textbf{Personalized FL.}
Instead of learning a common global model, personalized FL methods have been explored to maximize local performance by training personalized models for each client. Our framework can also be regarded as the personalized FL method by learning personalized BN layers. However, compared with existing personalized FL methods, which typically suffer a negative aspect of the lack of a common global model after training, our method provides a complete solution for testing on either internal or external clients, boosting the applicability.

\section{Conclusion}
This work proposes a novel and unified federated framework, UniFed, for tackling the non-iid features problem. Our proposed method promotes both federated training and testing with non-iid features via one unified solution, i.e., let each internal/external client use its specific batch normalization layers. We provide theoretical guarantees for a faster convergence speed and lower generalization error bound of client-specific BN compared with aggregated BN. We also perform causal reasoning for UniFed to provide a novel viewpoint to explain the effectiveness of our framework. In experiments, our extensive evaluations have demonstrated that UniFed can significantly improve the convergence behavior, model performance, as well as generalization capability on both natural and medical images with the presence of non-iid features. We also perform comprehensive analytical studies to investigate several key factors which may affect the model prediction for the real-world application.

We believe that UniFed can improve a wide range of applications in healthcare~\cite{rieke2020future}. A few interesting directions for future work include extending the validation in more scenarios such as autonomous driving~\cite{kamp2018efficient} and analyzing what types of differences in local data can benefit from UniFed.
Furthermore, it would be interesting to investigate the importance of position for BN layers in the network to dynamically adapt BN for internal/external clients.

\bibliographystyle{IEEEtran}
\bibliography{IEEEabrv,./ref}

\begin{thebibliography}{10}
\providecommand{\url}[1]{#1}
\csname url@samestyle\endcsname
\providecommand{\newblock}{\relax}
\providecommand{\bibinfo}[2]{#2}
\providecommand{\BIBentrySTDinterwordspacing}{\spaceskip=0pt\relax}
\providecommand{\BIBentryALTinterwordstretchfactor}{4}
\providecommand{\BIBentryALTinterwordspacing}{\spaceskip=\fontdimen2\font plus
\BIBentryALTinterwordstretchfactor\fontdimen3\font minus
  \fontdimen4\font\relax}
\providecommand{\BIBforeignlanguage}[2]{{%
\expandafter\ifx\csname l@#1\endcsname\relax
\typeout{** WARNING: IEEEtran.bst: No hyphenation pattern has been}%
\typeout{** loaded for the language `#1'. Using the pattern for}%
\typeout{** the default language instead.}%
\else
\language=\csname l@#1\endcsname
\fi
#2}}
\providecommand{\BIBdecl}{\relax}
\BIBdecl

\bibitem{rieke2020future}
N.~Rieke, J.~Hancox, W.~Li, F.~Milletari, H.~R. Roth, S.~Albarqouni, S.~Bakas,
  M.~N. Galtier, B.~A. Landman, K.~Maier-Hein \emph{et~al.}, ``The future of
  digital health with federated learning,'' \emph{NPJ digital medicine},
  vol.~3, no.~1, pp. 1--7, 2020.

\bibitem{wu2022communication}
C.~Wu, F.~Wu, L.~Lyu, Y.~Huang, and X.~Xie, ``Communication-efficient federated
  learning via knowledge distillation,'' \emph{Nature communications}, vol.~13,
  no.~1, pp. 1--8, 2022.

\bibitem{feddyn}
A.~E. Durmus, Z.~Yue, M.~Ramon, M.~Matthew, W.~Paul, and S.~Venkatesh,
  ``Federated learning based on dynamic regularization,'' in \emph{ICLR}, 2021.

\bibitem{sheller2020federated}
M.~J. Sheller, B.~Edwards, G.~A. Reina, J.~Martin, S.~Pati, A.~Kotrotsou,
  M.~Milchenko, W.~Xu, D.~Marcus \emph{et~al.}, ``Federated learning in
  medicine: facilitating multi-institutional collaborations without sharing
  patient data,'' \emph{Scientific reports}, vol.~10, no.~1, pp. 1--12, 2020.

\bibitem{dayan2021federated}
I.~Dayan, H.~R. Roth, A.~Zhong, A.~Harouni, A.~Gentili, A.~Z. Abidin, A.~Liu,
  A.~B. Costa, B.~J. Wood, C.-S. Tsai \emph{et~al.}, ``Federated learning for
  predicting clinical outcomes in patients with covid-19,'' \emph{Nature
  medicine}, vol.~27, no.~10, pp. 1735--1743, 2021.

\bibitem{dou2021federated}
Q.~Dou, T.~Y. So, M.~Jiang, Q.~Liu, V.~Vardhanabhuti, G.~Kaissis, Z.~Li, W.~Si,
  H.~H. Lee \emph{et~al.}, ``Federated deep learning for detecting covid-19
  lung abnormalities in ct: a privacy-preserving multinational validation
  study,'' \emph{NPJ digital medicine}, vol.~4, no.~1, pp. 1--11, 2021.

\bibitem{pati2022federated}
S.~Pati, U.~Baid, B.~Edwards, M.~Sheller, S.-H. Wang, G.~A. Reina, P.~Foley
  \emph{et~al.}, ``Federated learning enables big data for rare cancer boundary
  detection,'' \emph{arXiv preprint arXiv:2204.10836}, 2022.

\bibitem{fedavg}
B.~McMahan, E.~Moore, D.~Ramage, S.~Hampson, and B.~A. y~Arcas,
  ``Communication-efficient learning of deep networks from decentralized
  data,'' in \emph{AISTATS}, 2017, pp. 1273--1282.

\bibitem{hsieh2020non}
K.~Hsieh, A.~Phanishayee, O.~Mutlu, and P.~Gibbons, ``The non-iid data quagmire
  of decentralized machine learning,'' in \emph{ICML}.\hskip 1em plus 0.5em
  minus 0.4em\relax PMLR, 2020.

\bibitem{flchallenges}
T.~Li, A.~K. Sahu, A.~Talwalkar, and V.~Smith, ``Federated learning:
  Challenges, methods, and future directions,'' \emph{IEEE Signal Processing
  Magazine}, vol.~37, no.~3, pp. 50--60, 2020.

\bibitem{kairouz2021advances}
P.~Kairouz, H.~B. McMahan, B.~Avent, A.~Bellet, M.~Bennis, A.~N. Bhagoji,
  K.~Bonawitz, Z.~Charles, G.~Cormode, R.~Cummings \emph{et~al.}, ``Advances
  and open problems in federated learning,'' \emph{Foundations and
  Trends{\textregistered} in Machine Learning}, vol.~14, no. 1--2, pp. 1--210,
  2021.

\bibitem{li2019convergence}
X.~Li, K.~Huang, W.~Yang, S.~Wang, and Z.~Zhang, ``On the convergence of fedavg
  on non-iid data,'' in \emph{ICLR}, 2019.

\bibitem{scaffold}
S.~P. Karimireddy, S.~Kale, M.~Mohri, S.~Reddi, S.~Stich, and A.~T. Suresh,
  ``{SCAFFOLD}: Stochastic controlled averaging for federated learning,'' in
  \emph{ICML}, 2020.

\bibitem{liu2021feddg}
Q.~Liu, C.~Chen, J.~Qin, Q.~Dou, and P.-A. Heng, ``Feddg: Federated domain
  generalization on medical image segmentation via episodic learning in
  continuous frequency space,'' \emph{CVPR}, 2021.

\bibitem{fedadg}
L.~Zhang, X.~Lei, Y.~Shi, H.~Huang, and C.~Chen, ``Federated learning with
  domain generalization,'' \emph{arXiv:2111.10487}, 2021.

\bibitem{chen2022federated}
J.~Chen, M.~Jiang, Q.~Dou, and Q.~Chen, ``Federated domain generalization for
  image recognition via cross-client style transfer,'' \emph{arXiv preprint
  arXiv:2210.00912}, 2022.

\bibitem{aubreville2020completely}
M.~Aubreville, C.~A. Bertram, T.~A. Donovan, C.~Marzahl, A.~Maier, and
  R.~Klopfleisch, ``A completely annotated whole slide image dataset of canine
  breast cancer to aid human breast cancer research,'' \emph{Scientific data},
  vol.~7, no.~1, pp. 1--10, 2020.

\bibitem{liu2020ms}
Q.~Liu, Q.~Dou, L.~Yu, and P.~A. Heng, ``Ms-net: Multi-site network for
  improving prostate segmentation with heterogeneous mri data,'' \emph{IEEE
  Transactions on Medical Imaging}, 2020.

\bibitem{xu2021federated}
J.~Xu, B.~S. Glicksberg, C.~Su, P.~Walker, J.~Bian, and F.~Wang, ``Federated
  learning for healthcare informatics,'' \emph{Journal of Healthcare
  Informatics Research}, vol.~5, no.~1, pp. 1--19, 2021.

\bibitem{li2016revisiting}
Y.~Li, N.~Wang, J.~Shi, J.~Liu, and X.~Hou, ``Revisiting batch normalization
  for practical domain adaptation,'' \emph{arXiv preprint arXiv:1603.04779},
  2016.

\bibitem{huang2017arbitrary}
X.~Huang and S.~Belongie, ``Arbitrary style transfer in real-time with adaptive
  instance normalization,'' in \emph{ICCV}, 2017, pp. 1501--1510.

\bibitem{chang2019domain}
W.-G. Chang, T.~You, S.~Seo, S.~Kwak, and B.~Han, ``Domain-specific batch
  normalization for unsupervised domain adaptation,'' in \emph{CVPR}, 2019, pp.
  7354--7362.

\bibitem{choi2021robustnet}
S.~Choi, S.~Jung, H.~Yun, J.~T. Kim, S.~Kim, and J.~Choo, ``Robustnet:
  Improving domain generalization in urban-scene segmentation via instance
  selective whitening,'' in \emph{CVPR}, 2021, pp. 11\,580--11\,590.

\bibitem{ioffe2015batch}
S.~Ioffe and C.~Szegedy, ``Batch normalization: Accelerating deep network
  training by reducing internal covariate shift,'' in \emph{ICML}, 2015, pp.
  448--456.

\bibitem{santurkar2018does}
S.~Santurkar, D.~Tsipras, A.~Ilyas, and A.~Madry, ``How does batch
  normalization help optimization?'' in \emph{NeurIPS}, 2018, pp. 2483--2493.

\bibitem{benz2021revisiting}
P.~Benz, C.~Zhang, A.~Karjauv, and I.~S. Kweon, ``Revisiting batch
  normalization for improving corruption robustness,'' in \emph{WACV}, 2021,
  pp. 494--503.

\bibitem{fedbn}
X.~Li, M.~Jiang, X.~Zhang, M.~Kamp, and Q.~Dou, ``Fed{BN}: Federated learning
  on non-{IID} features via local batch normalization,'' in \emph{ICLR}, 2021.

\bibitem{jacot2018neural}
A.~Jacot, F.~Gabriel, and C.~Hongler, ``Neural tangent kernel: Convergence and
  generalization in neural networks,'' in \emph{NeurIPS}, 2018, pp. 8571--8580.

\bibitem{gopnik2004theory}
A.~Gopnik, C.~Glymour, D.~M. Sobel, L.~E. Schulz, T.~Kushnir, and D.~Danks, ``A
  theory of causal learning in children: causal maps and bayes nets.''
  \emph{Psychological review}, vol. 111, no.~1, p.~3, 2004.

\bibitem{geirhos2018generalisation}
R.~Geirhos, C.~R. Temme, J.~Rauber, H.~H. Sch{\"u}tt, M.~Bethge, and F.~A.
  Wichmann, ``Generalisation in humans and deep neural networks,''
  \emph{NeurIPS}, vol.~31, 2018.

\bibitem{pearl2009causality}
J.~Pearl, \emph{Causality}.\hskip 1em plus 0.5em minus 0.4em\relax Cambridge
  university press, 2009.

\bibitem{peters2017elements}
J.~Peters, D.~Janzing, and B.~Sch{\"o}lkopf, \emph{Elements of causal
  inference: foundations and learning algorithms}.\hskip 1em plus 0.5em minus
  0.4em\relax The MIT Press, 2017.

\bibitem{fedprox}
T.~Li, A.~K. Sahu, M.~Zaheer, M.~Sanjabi, A.~Talwalkar, and V.~Smith,
  ``Federated optimization in heterogeneous networks,'' in \emph{Conference on
  Machine Learning and Systems}, 2020.

\bibitem{moon}
Q.~Li, B.~He, and D.~Song, ``Model-contrastive federated learning,'' in
  \emph{CVPR}, 2021, pp. 10\,713--10\,722.

\bibitem{fedadam}
S.~J. Reddi, Z.~Charles, M.~Zaheer, Z.~Garrett, K.~Rush, J.~Kone{\v{c}}n{\'y},
  S.~Kumar, and H.~B. McMahan, ``Adaptive federated optimization,'' in
  \emph{ICLR}, 2021.

\bibitem{feddc}
L.~Gao, H.~Fu, L.~Li, Y.~Chen, M.~Xu, and C.-Z. Xu, ``Feddc: Federated learning
  with non-iid data via local drift decoupling and correction,'' in
  \emph{CVPR}, 2022.

\bibitem{fednova}
J.~Wang, Q.~Liu, H.~Liang, G.~Joshi, and H.~V. Poor, ``Tackling the objective
  inconsistency problem in heterogeneous federated optimization,''
  \emph{NeurIPS}, vol.~33, 2020.

\bibitem{pillutla2019robust}
K.~Pillutla, S.~M. Kakade, and Z.~Harchaoui, ``Robust aggregation for federated
  learning,'' \emph{arXiv preprint arXiv:1912.13445}, 2019.

\bibitem{reisizadeh2020robust}
A.~Reisizadeh, F.~Farnia, R.~Pedarsani, and A.~Jadbabaie, ``Robust federated
  learning: The case of affine distribution shifts,'' \emph{arXiv preprint
  arXiv:2006.08907}, 2020.

\bibitem{silobn}
M.~Andreux, J.~O. du~Terrail, C.~Beguier, and E.~W. Tramel, ``Siloed federated
  learning for multi-centric histopathology datasets,'' in
  \emph{DART-DCL}.\hskip 1em plus 0.5em minus 0.4em\relax Springer, 2020, pp.
  129--139.

\bibitem{perfedavg}
A.~Fallah, A.~Mokhtari, and A.~Ozdaglar, ``Personalized federated learning with
  theoretical guarantees: A model-agnostic meta-learning approach,''
  \emph{NeurIPS}, vol.~33, pp. 3557--3568, 2020.

\bibitem{li2021ditto}
T.~Li, S.~Hu, A.~Beirami, and V.~Smith, ``Ditto: Fair and robust federated
  learning through personalization,'' in \emph{ICML}.\hskip 1em plus 0.5em
  minus 0.4em\relax PMLR, 2021, pp. 6357--6368.

\bibitem{fedbabu}
J.~Oh, S.~Kim, and S.-Y. Yun, ``Fedbabu: Towards enhanced representation for
  federated image classification,'' \emph{arXiv preprint arXiv:2106.06042},
  2021.

\bibitem{yuan2022what}
H.~Yuan, W.~R. Morningstar, L.~Ning, and K.~Singhal, ``What do we mean by
  generalization in federated learning?'' in \emph{ICLR}, 2022.

\bibitem{fedsam}
Z.~Qu, X.~Li, R.~Duan, Y.~Liu, B.~Tang, and Z.~Lu, ``Generalized federated
  learning via sharpness aware minimization,'' in \emph{ICML}.\hskip 1em plus
  0.5em minus 0.4em\relax PMLR, 2022, pp. 18\,250--18\,280.

\bibitem{gauriau2021deep}
R.~Gauriau, B.~C. Bizzo, F.~C. Kitamura, O.~Landi~Junior, S.~F. Ferraciolli,
  F.~B. Macruz \emph{et~al.}, ``A deep learning--based model for detecting
  abnormalities on brain mr images for triaging: preliminary results from a
  multisite experience,'' \emph{Radiology: Artificial Intelligence}, vol.~3,
  no.~4, p. e200184, 2021.

\bibitem{li2021dynamic}
Y.~Li, L.~Yuan, Y.~Chen, P.~Wang, and N.~Vasconcelos, ``Dynamic transfer for
  multi-source domain adaptation,'' in \emph{CVPR}, 2021, pp. 10\,998--11\,007.

\bibitem{galstyan2022failure}
T.~Galstyan, H.~Harutyunyan, H.~Khachatrian, G.~V. Steeg, and A.~Galstyan,
  ``Failure modes of domain generalization algorithms,'' in \emph{CVPR}, 2022,
  pp. 19\,077--19\,086.

\bibitem{luo2018towards}
P.~Luo, X.~Wang, W.~Shao, and Z.~Peng, ``Towards understanding regularization
  in batch normalization,'' \emph{ICLR}, 2019.

\bibitem{yang2018a}
G.~Yang, J.~Pennington, V.~Rao, J.~Sohl-Dickstein, and S.~S. Schoenholz, ``A
  mean field theory of batch normalization,'' in \emph{ICLR}, 2019.

\bibitem{morcos2018importance}
A.~S. Morcos, D.~G. Barrett, N.~C. Rabinowitz, and M.~Botvinick, ``On the
  importance of single directions for generalization,'' \emph{arXiv preprint
  arXiv:1803.06959}, 2018.

\bibitem{li2018adaptive}
Y.~Li, N.~Wang, J.~Shi, X.~Hou, and J.~Liu, ``Adaptive batch normalization for
  practical domain adaptation,'' \emph{Pattern Recognition}, vol.~80, pp.
  109--117, 2018.

\bibitem{kohler2019exponential}
J.~Kohler, H.~Daneshmand, A.~Lucchi, T.~Hofmann, M.~Zhou, and K.~Neymeyr,
  ``Exponential convergence rates for batch normalization: The power of
  length-direction decoupling in non-convex optimization,'' in
  \emph{AISTATS}.\hskip 1em plus 0.5em minus 0.4em\relax PMLR, 2019, pp.
  806--815.

\bibitem{salimans2016weight}
T.~Salimans and D.~P. Kingma, ``Weight normalization: A simple
  reparameterization to accelerate training of deep neural networks,'' in
  \emph{NeurIPS}, 2016, pp. 901--909.

\bibitem{arora2019fine}
S.~Arora, S.~S. Du, W.~Hu, Z.~Li, and R.~Wang, ``Fine-grained analysis of
  optimization and generalization for overparameterized two-layer neural
  networks,'' \emph{arXiv preprint arXiv:1901.08584}, 2019.

\bibitem{du2018gradient}
S.~S. Du, X.~Zhai, B.~Poczos, and A.~Singh, ``Gradient descent provably
  optimizes over-parameterized neural networks,'' \emph{arXiv preprint
  arXiv:1810.02054}, 2018.

\bibitem{allen2019convergence}
Z.~Allen-Zhu, Y.~Li, and Z.~Song, ``A convergence theory for deep learning via
  over-parameterization,'' in \emph{ICML}.\hskip 1em plus 0.5em minus
  0.4em\relax PMLR, 2019, pp. 242--252.

\bibitem{van2020training}
J.~van~den Brand, B.~Peng, Z.~Song, and O.~Weinstein, ``Training
  (overparametrized) neural networks in near-linear time,'' \emph{arXiv
  e-prints}, pp. arXiv--2006, 2020.

\bibitem{dukler2020optimization}
Y.~Dukler, Q.~Gu, and G.~Montúfar, ``Optimization theory for relu neural
  networks trained with normalization layers,'' 2020.

\bibitem{blitzer2007learning}
J.~Blitzer, K.~Crammer, A.~Kulesza, F.~Pereira, and J.~Wortman, ``Learning
  bounds for domain adaptation,'' in \emph{NeurIPS}, 2007, pp. 129--136.

\bibitem{albuquerque2019generalizing}
I.~Albuquerque, J.~Monteiro, M.~Darvishi, T.~H. Falk, and I.~Mitliagkas,
  ``Generalizing to unseen domains via distribution matching,'' \emph{arXiv
  preprint arXiv:1911.00804}, 2019.

\bibitem{dodge2017study}
S.~Dodge and L.~Karam, ``A study and comparison of human and deep learning
  recognition performance under visual distortions,'' in \emph{ICCCN}.\hskip
  1em plus 0.5em minus 0.4em\relax IEEE, 2017, pp. 1--7.

\bibitem{pearl2016causal}
J.~Pearl, M.~Glymour, and N.~P. Jewell, \emph{Causal inference in statistics: A
  primer}.\hskip 1em plus 0.5em minus 0.4em\relax John Wiley \& Sons, 2016.

\bibitem{arjovsky2019invariant}
M.~Arjovsky, L.~Bottou, I.~Gulrajani, and D.~Lopez-Paz, ``Invariant risk
  minimization,'' \emph{arXiv preprint arXiv:1907.02893}, 2019.

\bibitem{piratla2020efficient}
V.~Piratla, P.~Netrapalli, and S.~Sarawagi, ``Efficient domain generalization
  via common-specific low-rank decomposition,'' in \emph{ICML}, 2020, pp.
  7728--7738.

\bibitem{zhao2020domain}
S.~Zhao, M.~Gong, T.~Liu, H.~Fu, and D.~Tao, ``Domain generalization via
  entropy regularization,'' \emph{NeurIPS}, vol.~33, 2020.

\bibitem{zhao2018adversarial}
H.~Zhao, S.~Zhang, G.~Wu, J.~M. Moura, J.~P. Costeira, and G.~J. Gordon,
  ``Adversarial multiple source domain adaptation,'' \emph{NeurIPS}, vol.~31,
  pp. 8559--8570, 2018.

\bibitem{li2018federated}
T.~Li, A.~K. Sahu, M.~Zaheer, M.~Sanjabi, A.~Talwalkar, and V.~Smith,
  ``Federated optimization in heterogeneous networks,'' in \emph{Conference on
  Machine Learning and Systems}, 2020.

\bibitem{reddi2021adaptive}
S.~J. Reddi, Z.~Charles, M.~Zaheer, Z.~Garrett, K.~Rush, J.~Kone{\v{c}}n{\'y},
  S.~Kumar, and H.~B. McMahan, ``Adaptive federated optimization,'' in
  \emph{ICLR}, 2021.

\bibitem{wang2020tackling}
J.~Wang, Q.~Liu, H.~Liang, G.~Joshi, and H.~V. Poor, ``Tackling the objective
  inconsistency problem in heterogeneous federated optimization,''
  \emph{NeurIPS}, vol.~33, 2020.

\bibitem{li2021fedbn}
X.~Li, M.~Jiang, X.~Zhang, M.~Kamp, and Q.~Dou, ``Fed{BN}: Federated learning
  on non-{IID} features via local batch normalization,'' in \emph{ICLR}, 2021.

\bibitem{hendrycks2019robustness}
D.~Hendrycks and T.~Dietterich, ``Benchmarking neural network robustness to
  common corruptions and perturbations,'' \emph{ICLR}, 2019.

\bibitem{krizhevsky2012imagenet}
A.~Krizhevsky, I.~Sutskever, and G.~E. Hinton, ``Imagenet classification with
  deep convolutional neural networks,'' \emph{NeurIPS}, vol.~25, pp.
  1097--1105, 2012.

\bibitem{bandi2018detection}
P.~Bandi, O.~Geessink, Q.~Manson, M.~Van~Dijk, M.~Balkenhol, M.~Hermsen, B.~E.
  Bejnordi, B.~Lee, K.~Paeng, A.~Zhong \emph{et~al.}, ``From detection of
  individual metastases to classification of lymph node status at the patient
  level: the camelyon17 challenge,'' \emph{IEEE Transactions on Medical
  Imaging}, 2018.

\bibitem{koh2020wilds}
P.~W. Koh, S.~Sagawa, H.~Marklund, S.~M. Xie, M.~Zhang, A.~Balsubramani, W.~Hu,
  M.~Yasunaga, R.~L. Phillips, I.~Gao \emph{et~al.}, ``Wilds: A benchmark of
  in-the-wild distribution shifts,'' \emph{arXiv preprint arXiv:2012.07421},
  2020.

\bibitem{huang2017densely}
G.~Huang, Z.~Liu, L.~Van Der~Maaten, and K.~Q. Weinberger, ``Densely connected
  convolutional networks,'' in \emph{CVPR}, 2017, pp. 4700--4708.

\bibitem{liu2022medical}
X.~Liu, B.~Glocker, M.~M. McCradden, M.~Ghassemi, A.~K. Denniston, and
  L.~Oakden-Rayner, ``The medical algorithmic audit,'' \emph{The Lancet Digital
  Health}, 2022.

\bibitem{geiping2020inverting}
J.~Geiping, H.~Bauermeister, H.~Dr{\"o}ge, and M.~Moeller, ``Inverting
  gradients--how easy is it to break privacy in federated learning?''
  \emph{arXiv preprint arXiv:2003.14053}, 2020.

\bibitem{hatamizadeh2022gradient}
A.~Hatamizadeh, H.~Yin, P.~Molchanov, A.~Myronenko, W.~Li, P.~Dogra, A.~Feng,
  M.~G. Flores, J.~Kautz, D.~Xu \emph{et~al.}, ``Do gradient inversion attacks
  make federated learning unsafe?'' \emph{arXiv preprint arXiv:2202.06924},
  2022.

\bibitem{huang2021evaluating}
Y.~Huang, S.~Gupta, Z.~Song, K.~Li, and S.~Arora, ``Evaluating gradient
  inversion attacks and defenses in federated learning,'' \emph{NeurIPS},
  vol.~34, pp. 7232--7241, 2021.

\bibitem{kamp2018efficient}
M.~Kamp, L.~Adilova, J.~Sicking, F.~H{\"u}ger, P.~Schlicht, T.~Wirtz, and
  S.~Wrobel, ``Efficient decentralized deep learning by dynamic model
  averaging,'' in \emph{Joint European conference on machine learning and
  knowledge discovery in databases}.\hskip 1em plus 0.5em minus 0.4em\relax
  Springer, 2018, pp. 393--409.

\end{thebibliography}

\begin{IEEEbiography}[{\includegraphics[width=1in,height=1.25in,clip,keepaspectratio]{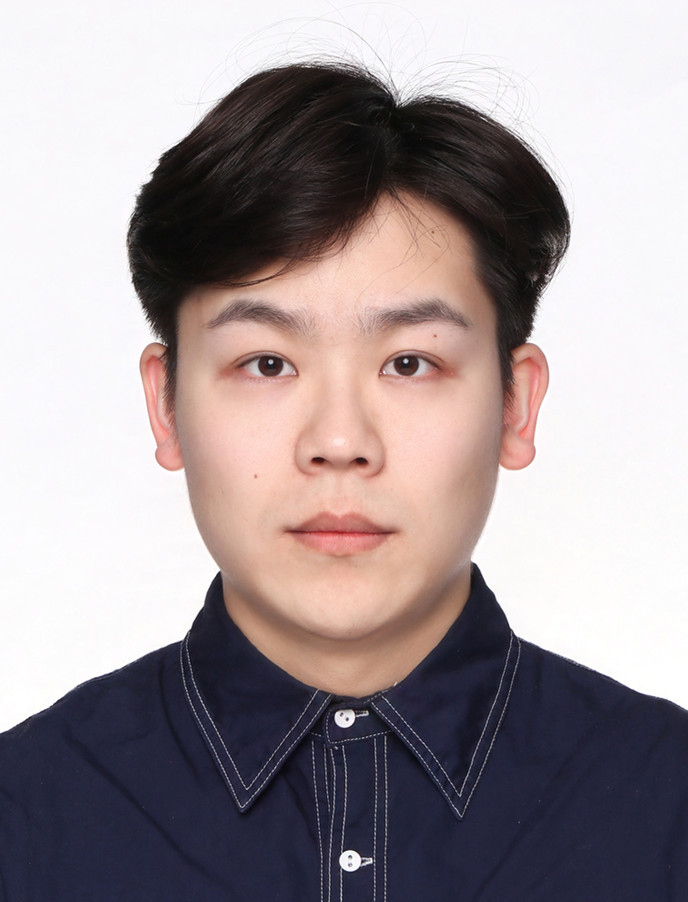}}]{Meirui Jiang} received the B.Eng degree in Shandong University, China, in 2020. He is currently working toward the Ph.D. degree with the Department of Computer Science and Engineering, The Chinese University of Hong Kong.
His research interests include medical image analysis and federated learning.
\end{IEEEbiography}

\begin{IEEEbiography}[{\includegraphics[width=1in,height=1.25in,clip,keepaspectratio]{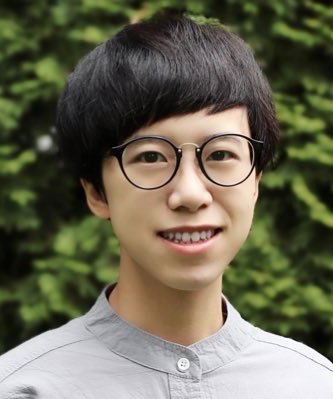}}]{Xiaoxiao Li} is an Assistant Professor at the Department of Electrical and Computer Engineering (ECE) at the University of British Columbia (UBC) starting August 2021. Before joining UBC, Dr. Li was a Postdoc Research Fellow in the Computer Science Department at Princeton University. Dr. Li obtained her PhD degree from Yale University in 2020. Dr. Li’s research interests range across the interdisciplinary fields of deep learning and biomedical data analysis, aiming to improve the trustworthiness of AI systems for healthcare. In the last few years, Dr. Li has had over 30 papers published in leading machine learning conferences and journals, including NeurIPS, ICML, ICLR, MICCAI, IPMI, BMVC, IEEE Transactions on Medical Imaging, and Medical Image Analysis. Her work has been recognized with several best paper awards at international conferences.
\end{IEEEbiography}

\begin{IEEEbiography}[{\includegraphics[width=1in,height=1.25in,clip,keepaspectratio]{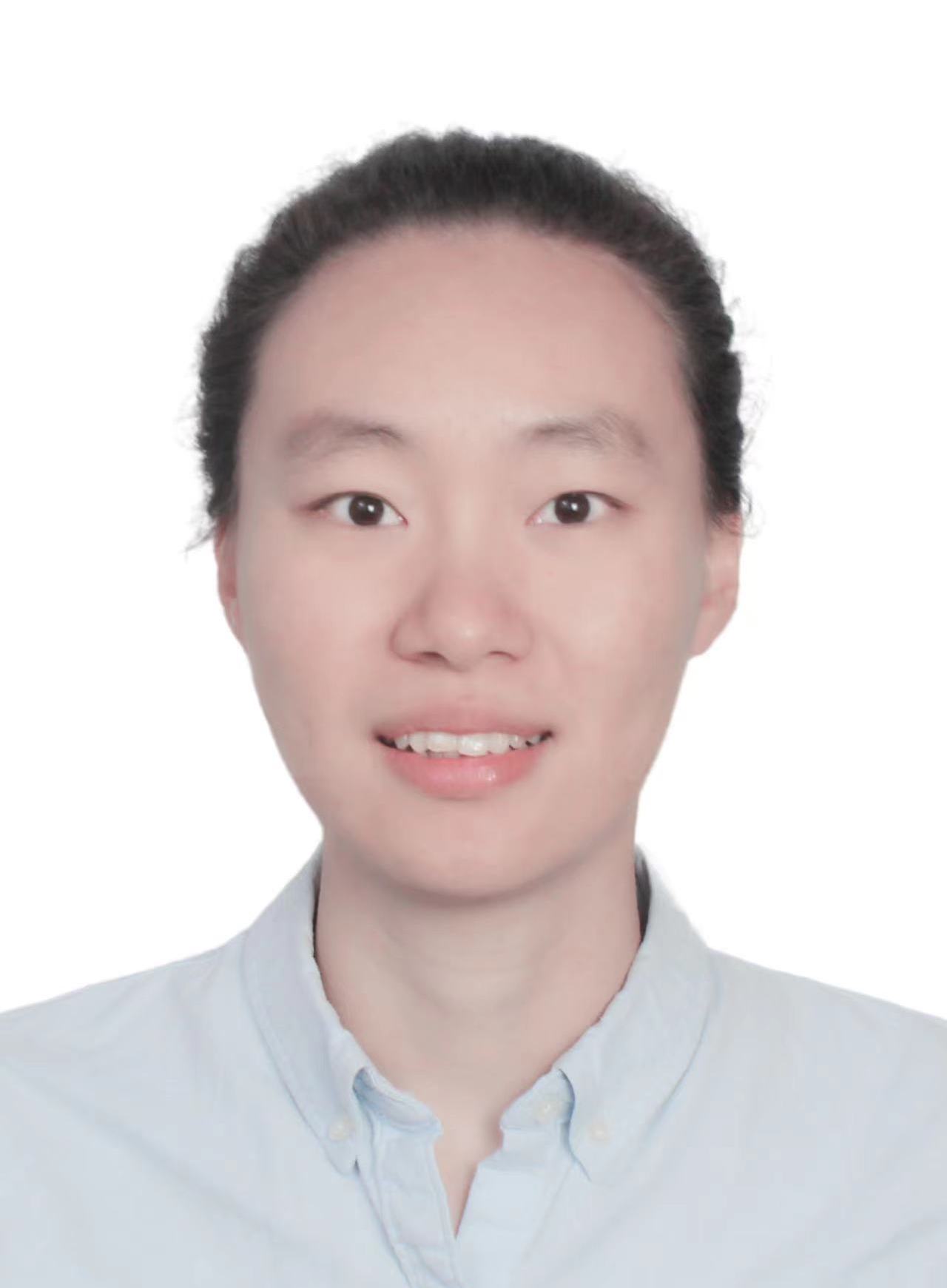}}]{Xiaofei Zhang} received the Ph.D. degree in Department of Statistics from Iowa State University in 2021. She is an assistant professor with School of Statistics and Mathematics, at Zhongnan University of Economics and Law.
\end{IEEEbiography}

\begin{IEEEbiography}[{\includegraphics[width=1in,height=1.25in,clip,keepaspectratio]{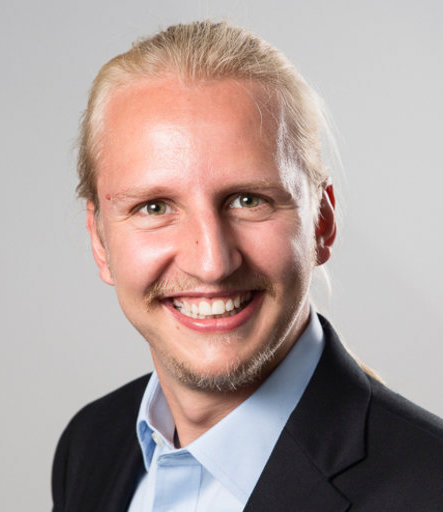}}]{Michael Kamp} received the Ph.D. degree in Computer Science from University of Bonn. He is a leader of the research group Trustworthy Machine Learning at the Institut für KI in der Medizin (IKIM), located at the Ruhr-University Bochum. His research interests lie in the theoretically sound application of machine learning to distributed data sources which entails four major challenges: The computational complexity of processing very large datasets, the often prohibitive communication required to centralize this data, the privacy-issues of sharing highly sensitive data, and the trustworthiness of the resulting model. 
\end{IEEEbiography}

\begin{IEEEbiography}[{\includegraphics[width=1in,height=1.25in,clip,keepaspectratio]{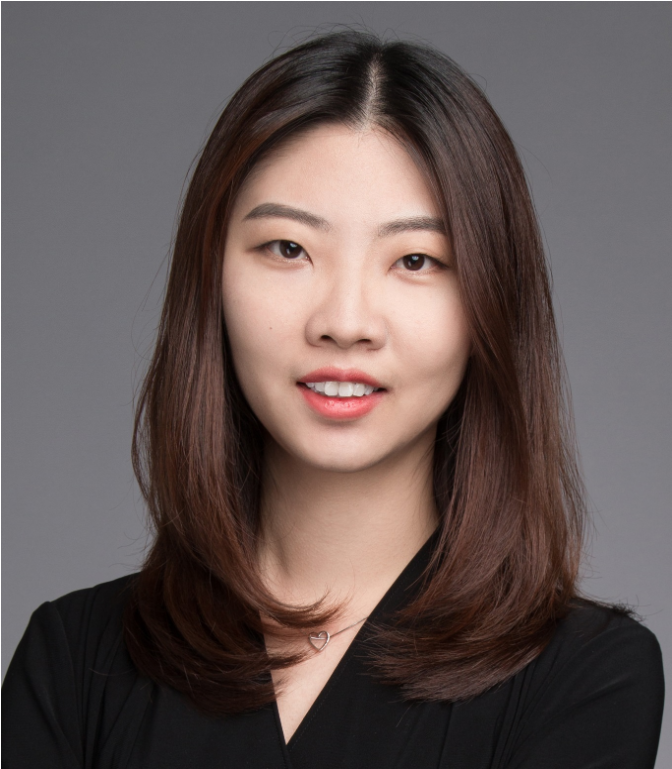}}]{Qi Dou} received the Ph.D. degree in Computer Science and Engineering at The Chinese University of Hong Kong in 2018, and B.Eng degree in Biomedical Engineering at Beihang University in 2014. She is an assistant professor with the Department of Computer Science and Engineering, and co-affiliated with T Stone Robotics Institute, at The Chinese University of Hong Kong. Her research interest lies in the interdisciplinary area of artificial intelligence and healthcare with expertise in medical image analysis and robotic surgery, with the mission to advance disease diagnosis and minimally invasive intervention via machine intelligence. 
\end{IEEEbiography}

\clearpage
\appendices

\textbf{Roadmap of Appendix} The Appendix is organized as follows.
We provide theoretical proof of convergence in Section~\ref{app:convergence_proof}. 
We provide theoretical proof of generalization error bound in Section~\ref{app:error_bound}. And we provide the explanations and proofs for causal reasoning in Section~\ref{app:causal_analysis_proof}.

\section{Proof of Convergence}
\label{app:convergence_proof}
\subsection{Evolution Dynamics}
\label{sec:evol}
In this section, we calculate the evolution dynamics $\boldsymbol{\Lambda}(t)$ for training with function $f$ and $\boldsymbol{\Lambda}^*(t)$ for training with $f^*$. 
Since the parameters are updated using gradient descent, the optimization dynamics of parameters are
\begin{align*}
\frac{d \mathbf{v}_{k}}{d t}=-\frac{\partial L}{\partial \mathbf{v}_{k}}, \quad \frac{d \gamma_{k}}{d t} =-\frac{\partial L}{\partial \gamma_{k}}.
\end{align*}
Let $f_p = f(\b x_p^{i_p})$. Then, the dynamics of the prediction of the $p$-th data point in site $i_p$ is 
\begin{align*}
\frac{\partial f_{p}}{\partial t}
&=\sum_{k=1}^{m} \frac{\partial f_{p}}{\partial \mathbf{v}_{k}} \frac{d \mathbf{v}_{k}}{d t}+\frac{\partial f_{p}}{\partial \gamma_{k}} \frac{d \gamma_{k}}{d t}\\
&=-\underbrace{\sum_{k=1}^{m} \frac{\partial f_{p}}{\partial \mathbf{v}_{k}} \frac{\partial L}{\partial \mathbf{v}_{k}}}_{T_{\mathbf{v}}^{p}}-\underbrace{\sum_{k=1}^{m} \frac{\partial f_{p}}{\partial \gamma_{k}} \frac{\partial L}{\partial \gamma_{k}}}_{T_{\gamma}^{p}}.
\end{align*}
The gradients of $f_p$ and $L$ with respect to $\b v_k$ and $\gamma_k$ are computed as
\begin{align*}
\begin{aligned}
\frac{\partial f_{p}}{\partial \mathbf{v}_{k}}(t) &=\frac{1}{\sqrt{m}} \frac{c_{k} \cdot \gamma_{k}(t)}{\left\|\mathbf{v}_{k}(t)\right\|_{\b S_{i_p}}} \cdot \b x_{p}^{\mathbf{v}^{i_p}_{k}(t)^{\perp}} \bm{1}_{p k}(t), \\
\frac{\partial L}{\partial \mathbf{v}_{k}}(t) &=\frac{1}{\sqrt{m}} \sum_{q=1}^{N M}\left(f_{q}(t)-y_{q}\right) \frac{c_{k} \cdot \gamma_{k}(t)}{\left\|\mathbf{v}_{k}(t)\right\|_{\b S_{i_q}}} \b x_{q}^{\mathbf{v}^{i_q}_{k}(t)^{\perp}} \bm{1}_{q k}(t), \\
\frac{\partial f_{p}}{\partial \gamma_{k}}(t) &=\frac{1}{\sqrt{m}} \frac{c_{k}}{\left\|\mathbf{v}_{k}(t)\right\|_{\b S_{i_p}}} \phi\left(\mathbf{v}_{k}(t)^{\top} \mathbf{x}_{p}\right),  \\
\frac{\partial L}{\partial \gamma_{k}}(t) &=\frac{1}{\sqrt{m}} \sum_{q=1}^{N M}\left(f_{q}(t)-y_{q}\right) \frac{c_{k}}{\left\|\mathbf{v}_{k}(t)\right\|_{\b S_{i_q}}} \phi\left(\mathbf{v}_{k}(t)^{\top} \mathbf{x}_{q}\right),
\end{aligned}
\end{align*}
where $f_p = f(\b x^{i_p}_p)$, $\b x_{p}^{\mathbf{v}^{i_p}_{k}(t)^{\perp}} \triangleq (\b I - \frac{\b S_{i_p} \b u \b{u}^\top }{\parallel \b u \parallel_{\b S_{i_p}}^2})\b x$, and $\bm{1}_{pk}(t) \triangleq \bm{1}_{\{\b v_k(t)^\top \b x_p \geq 0 \}}.$
\\\\
We define Gram matrix $\b V(t)$ and $\b G(t)$ as
\begin{align}
\mathbf{V}_{p q}(t) & = 
\frac{1}{m} \sum_{k=1}^{m}\left(\alpha c_{k} \cdot \gamma_{k}(t)\right)^2 \left\|\mathbf{v}_{k}(t)\right\|_{\b S_{i_p}}^{-1} \left\|\mathbf{v}_{k}(t)\right\|_{\b S_{i_q}}^{-1} \notag 
\\&\quad\times\left\langle\mathbf{x}_{p}^{\mathbf{v}^{i_p}_{k}(t)^{\perp}}, \mathbf{x}_{q}^{\mathbf{v}^{i_q}_{k}(t)^{\perp}}\right\rangle \bm{1}_{p k}(t) \bm{1}_{q k}(t), \label{equ: fedavg_v} \\
\mathbf{G}_{p q}(t) & =
\frac{1}{m} \sum_{k=1}^{m}  c_{k}^{2} \left\|\mathbf{v}_{k}(t)\right\|_{\b S_{i_p}}^{-1} \left\|\mathbf{v}_{k}(t)\right\|_{\b S_{i_q}}^{-1} \notag
\\&\quad\times\phi\left(\mathbf{v}_{k}(t)^{\top} \mathbf{x}_{p}\right) \phi\left(\mathbf{v}_{k}(t)^{\top} \mathbf{x}_{q}\right). \label{equ: fedavg_g_apdx}
\end{align}

It follows that
\begin{align*}
T_{\mathbf{v}}^{p}(t)=\sum_{q=1}^{N M} \frac{\mathbf{V}_{p q}(t)}{\alpha^{2}}\left(f_{q}(t)-y_{q}\right), \\
T_{\mathbf{\gamma}}^{p}(t)=\sum_{q=1}^{N M} \mathbf{G}_{p q}(t)\left(f_{q}(t)-y_{q}\right).
\end{align*}
Let $\mathbf{f}=\left(f_{1}, \ldots, f_{n}\right)^{\top}=\left(f\left(\mathbf{x}_{1}\right), \ldots, f\left(\mathbf{x}_{NM}\right)\right)^{\top}$. The full evolution dynamic is given by
\begin{align*}
\frac{d \mathbf{f}}{d t}=-\boldsymbol{\Lambda}(t)(\mathbf{f}(t)- \b y), \quad \text{where} \quad
\boldsymbol{\Lambda}(t):=\frac{\mathbf{V}(t)}{\alpha^{2}}+\mathbf{G}(t).
\end{align*}

Similarly, we compute Gram matrix $\b V^*(t)$ and $\b G^*(t)$ for UniFed with $f^*$ as

\begin{align}
\mathbf{V}^*_{p q}(t) &= 
\frac{1}{m} \sum_{k=1}^{m}\left(\alpha c_{k} \right)^2 \gamma_{k,i_p}(t) \gamma_{k,i_q}(t) \left\|\mathbf{v}_{k}(t)\right\|_{\b S_{i_p}}^{-1} \left\|\mathbf{v}_{k}(t)\right\|_{\b S_{i_q}}^{-1}\notag
\\&\quad\times\left\langle\mathbf{x}_{p}^{\mathbf{v}^{i_p}_{k}(t)^{\perp}}, \mathbf{x}_{q}^{\mathbf{v}^{i_q}_{k}(t)^{\perp}}\right\rangle \bm{1}_{p k}(t) \bm{1}_{q k}(t), \label{equ: fedbn_v} \\
\mathbf{G}^*_{p q}(t) & =\frac{1}{m} \sum_{k=1}^{m}  c_{k}^{2} \left\|\mathbf{v}_{k}(t)\right\|_{\b S_{i_p}}^{-1} \left\|\mathbf{v}_{k}(t)\right\|_{\b S_{i_q}}^{-1}  \phi\left(\mathbf{v}_{k}(t)^{\top} \mathbf{x}_{p}\right)\notag \\&\quad\times\phi\left(\mathbf{v}_{k}(t)^{\top} \mathbf{x}_{q}\right) \bm{1}\{i_p = i_q\}. \label{equ: fedbn_g_apdx} 
\end{align}

Thus, the full evolution dynamic of UniFed is
\begin{align*}
\frac{d \mathbf{f}^*}{d t}=-\boldsymbol{\Lambda}^*(t)(\mathbf{f}^*(t)- \b y), \text{where} \hspace{0.5em}
\boldsymbol{\Lambda}^*(t):=\frac{\mathbf{V}^*(t)}{\alpha^{2}}+\mathbf{G}^*(t).
\end{align*}

\subsection{Proof of Lemma \ref{lemma:eigen}}
\label{sec:proof_lemma}
\cite{dukler2020optimization} proved that the matrix $\b G^{\infty}$ is strictly positive definite. In their proof, $\b G^{\infty}$ is the covariance matrix of the functionals $\psi_p$ define as
\[
\psi_{p}(\b v) := \phi \left( \mathbf{v}^{\top} \mathbf{x}_{p} \right) 
\]
over the Hilbert space $\mathcal{V}$ of $L^{2}\left(N\left(0, \alpha^{2} \mathbf{I}\right)\right)$. $\b G^{*\infty}$ is strictly positive definite by showing that $\psi_1,\cdots,\psi_{NM}$ are linearly independent, which is equivalent to that 
\begin{align}
\label{equ: inde_phi}
c_{1} \psi_{1}+c_{2} \psi_{2}+\cdots+c_{NM} \psi_{NM}=0 \text{ in } \mathcal{V}
\end{align}
holds only for $c_p=0$ for all $p$.

Let $\b G^{\infty}_i$ denote the $i$-th $M \times M$ block matrices on the diagonal of $\b G^{\infty}$. Then we have
\begin{align*}
    \b G^{*\infty} = diag(\b G^{\infty}_1,\cdots,\b G^{\infty}_N).
\end{align*}
To prove that $\b G^{*\infty}$ is strictly positive definite, we will show that $\b G^{\infty}_i$ is positive definite. Let us define 
\[
\psi^*_{j,i}(\b v) := \phi \left( \mathbf{v}^{\top} \mathbf{x}_{j} \right) \bm{1}\{j \in \text{ site } i \}, \quad j=1,\cdots, M.
\]
Then, we are going to show that
\begin{align}
\label{equ: inde_phi2}
c_{1} \psi^*_{1,i}+c_{2} \psi^*_{2,i}+\cdots+c_{M} \psi^*_{M,i}=0 
\end{align}
holds only for $c_j=0, \forall j \in [M]$. Suppose there exist
$c_{1},\cdots,c_{M}$ that are not identically 0, satisfying (\ref{equ: inde_phi2}). Let the coefficients for client $i$ be $c_{1},\cdots,c_{M}$ and let the coefficients for other client
be 0. Then, we have a sequence of coefficients satisfying (\ref{equ: inde_phi}), which is a contradiction with that $\b G^{\infty}$ is strictly positive definite. This implies $\b G^{\infty}_i$ is strictly positive definite. Namely, $\b G^{\infty}_i$'s eigenvalues are positive.  Since the eigenvalues of $\b G^{*\infty}$ are exactly the union of the eigenvalues of $\b G^{\infty}_i$, $\lambda_{min}(\b G^{*\infty})$ is positive and thus, $\b G^{*\infty}$ is strictly positive definite.

\subsection{Proof of Corollary \ref{coro:comp}}

To compare the convergence rates of using aggregated BN and client-specific BN, we compare the exponential factor in the convergence rates, which are $\left(1-\eta e_{0}/2\right)$ and $\left(1-\eta e^*_{0}/2\right)$, respectively.
Then, it reduces to comparing $e_{0} = \lambda_{\min}(\b G^{\infty})$ and $e^*_0 = \lambda_{\min}(\b G^{*\infty})$. 
Comparing equation (\ref{equ: fedavg_g_apdx}) and (\ref{equ: fedbn_g_apdx}), $\b G^{*\infty}$ takes the $M \times M$ block matrices on the diagonal of $\b G^{\infty}$:

\[\b G^{\infty}=\left[\begin{array}{cccc}
\b G^{\infty}_1 & \b G^{\infty}_{1,2}  & \cdots & \b G^{\infty}_{1,N}  \\
\b G^{\infty}_{1,2} & \b G^{\infty}_2 & \cdots & \b G^{\infty}_{2,N}  \\
\vdots & \vdots & \ddots & \vdots \\
\b G^{\infty}_{1,N} &\b G^{\infty}_{2,N} & \cdots & \b G^{\infty}_N
\end{array}\right], \]
\[\b G^{*\infty}=\left[\begin{array}{cccc}
\b G^{\infty}_{1} & 0 & \cdots & 0 \\
0 & \b G^{\infty}_{2} & \cdots & 0 \\
\vdots & \vdots & \ddots & \vdots \\
0 & 0 & \cdots &\b G^{\infty}_{N} \notag
\end{array}\right],\]

where $\b G^{\infty}_i$ is the $i$-th $M \times M$ block matrices on the diagonal of $\b G^{\infty}$. By linear algebra, 
\[
\lambda_{\min}(\b G^{\infty}_i) \geq \lambda_{\min}(\b G^{\infty}), \quad  \forall i \in [N].
\]
Since the eigenvalues of $\b G^{*\infty}$ are exactly the union of eigenvalues of $\b G^{\infty}_i$, we have 
\begin{align*}
\lambda_{\min}(\b G^{*\infty}) & = \min_{i \in [N]} \{ \lambda_{\min}(\b G^{\infty}_i) \}, \\
& \geq \lambda_{\min}(\b G^{\infty}).
\end{align*}
Thus, $\left(1-\eta e_{0}/2\right) \geq \left(1-\eta e^*_{0}/2\right)$ and we can conclude that the convergence rate of using client-specific BN is faster than the convergence of using aggregated BN.

\section{Error Bound Analysis and Proofs}
\label{app:error_bound}
In this section, we first give the proofs for the Theorem~\ref{thm:initial_bound}, then we extend this bound for federated learning with multiple training clients, and at last, we introduce a corollary showing the new error bound with reduced divergence term by using our client-specific BN.

\subsection{Proof of Theorem~\ref{thm:initial_bound} and Corollary~\ref{cor:error_bound_fl}}
\label{sec:proof_error_bound}
First, we introduce notation for our error bound analysis on out-of-federation clients.  

\textbf{Notation:}
To study the generalization error, we consider two types of clients, \textit{seen} clients (i.e., the internal training clients) and \textit{unseen} clients (i.e., the external testing clients). The distribution for seen and unseen clients are denoted as $\D^s$ and $\D^u$ respectively.

Moreover, we define a \emph{hypothesis} $f: \X \rightarrow \Y$, such that $f \in \cH$, where $\cH$ is a set of candidate hypothesis, and finally define the risk of hypothesis $f$ on distribution $\D^s$ as $J_s(f) = \E_{\b x \sim \D^s}[|f(\b x) - y|]$. We denote the risk of hypothesis $f$ on $\D^s$ as $J_s(f)$. Parallelly, the risk of hypothesis $f$ on $\D^u$ are denoted as $J_u(f)$.

\textbf{Distance between two distributions:} Let $\cH$ be a hypothesis class for input space $\X$, and $\A_{\cH}$ be the set of subsets of $\X$ that are the support of some hypothesis in $\cH$, i.e. for each $f \in \cH$, $\{\b x:\b x \in \X,f(\b x) = 1\} \in \A_{\cH}$. Then the distance between two distributions $\D$ and $\D'$ is defined as : $d_{\cH}(\D,\D'):=2 \sup _{A\in \A_{\cH}}|P_{\D}(A) - P_{\D'}(A)|$.

The symmetric difference space $\cH \Delta \cH$ is defined as : $\cH \Delta \cH:=\{f(\b x) \oplus f'(\b x) | f,f' \in \cH\}$, where $\oplus$ is the XOR operation. $\A_{\cH \Delta \cH}$ is defined as the set of all sets $A$ such that $A = \{\b x : \b x \in \X, f(\b x) \neq f'(\b x)\}$ for some $f,f' \in \cH$ and it follows that a distance $d_{\cH \Delta \cH}$ is well-defined. The \textit{ideal hypothesis} $f^o$ minimizes combined source and target risk: $f^o = \argmin_{f \in \cH} J_s(f)+J_u(f)$ and the error of $J^o$ is defined as $\epsilon := J_s(f^o)+J_u(f^o)$.

Based on \cite{albuquerque2019generalizing}, the error bound of the unseen client given one seen client is bounded by:

\begin{equation}
\label{app:initial_bound}
\begin{aligned}
    J_{u}(f) &\leq \hat{J}_{s}(f)+\frac{1}{2} \hat{d}_{\mathcal{H} \Delta \mathcal{H}}\left(\hat{\D^s}, \hat{\D^u}\right)
    \\&\quad+4 \sqrt{\frac{2 d \log \left(2M\right) +\log \left({4}/{\delta}\right)}{M}}+\epsilon.
\end{aligned}
\end{equation}

Consider federated learning with $N$ seen clients and an unseen client, samples size of each client is $M$, the upper bound of ${d}_{\mathcal{H} \Delta \mathcal{H}}$ can be further derived as below:
\begin{align*}
\begin{aligned}
{d}_{\cH \Delta \cH}\left({\mathcal{D}}^{{s}}, {\mathcal{D}}^{u}\right)
&=2 \sup _{A \in \mathcal{A}_{\mathcal{H} \Delta \mathcal{H}}}\left|{P}_{{\mathcal{D}}_{{S}}}(A)-P_{{\mathcal{D}}_{U}}(A)\right|
\\&=2 \sup _{A \in \mathcal{A}_{\mathcal{H} \Delta \mathcal{H}}}\left|\sum_{i=1}^N \omega_{i}\left(P_{{\mathcal{D}}^{s}_{i}}(A)-P_{{\mathcal{D}}^{u}}(A)\right)\right|
\\ &\leq 2 \sup _{A \in \mathcal{A}_{\mathcal{H} \Delta \mathcal{H}}} \sum_{i=1}^N \omega_{i}\left(\mid P_{{\mathcal{D}}^{s}_{i}}(A)-P_{{\mathcal{D}}^{u}}(A)\mid\right) 
\\ &\leq 2 \sum_{i=1}^N \omega_{i} \sup _{A \in \mathcal{A}_{\mathcal{H} \Delta \mathcal{H}}}\left(\mid P_{{\mathcal{D}}^{s}_{i}}(A)-{P}_{{\mathcal{D}}^{u}}(A)\mid\right)
\\ &=\sum_{i=1}^N \omega_{i} {d}_{\mathcal{H} \Delta \mathcal{H}}\left({\mathcal{D}}^s_{i}, {\D}^{u}\right),
\end{aligned}
\end{align*}
the first inequality is derived by the triangle inequality. With the same triangle inequality, we can get $\epsilon \leq \sum_{i=1}^N \epsilon_i$ similarly.
Note that ${J}_{s}(f_u) = {J}_{s}(\sum_{i=1}^N \omega_i f_{s,i})$ for $\forall f_u \in \cH$.
Replace the $\epsilon, {J}_{s}(f_u)$ and ${d}_{\mathcal{H} \Delta \mathcal{H}}$ into Eq.(\ref{app:initial_bound}), we have:

\vspace{-2mm}
\begin{equation}
\label{app:fl_bound}
\begin{aligned}
    J_{u}(f_{u}) &\leq \hat{J}_{\tilde{s}}(\sum_{i=1}^N\!\omega_i f_{s,i})\!+\!\!\sum_{i=1}^N \omega_i \!\!\left(\frac{1}{2} \hat{d}_{\mathcal{H} \Delta \mathcal{H}}\!\left(\hat{\D_i^s},\!\hat{\D^{u}}\right)\!+\!\epsilon_i\!\right)
    \\&\quad+4 \sqrt{\frac{2 d \log (2NM)+\log (4 / \delta)}{NM}}.
\end{aligned}
\end{equation}

With the assumption that the input data points follow Gaussian distribution for all clients, we denote the Gaussian distribution for client $i$ as $\N_i$, the Gaussian distribution for unseen client as $\N_u$ and the mixture of training samples with size $NM$ as $\tilde{\N}$ , we can change Eq.(~\ref{app:fl_bound}) into following form:
\begin{equation}
\label{app:fl_guassian_bound}
\begin{aligned}
J_{u}(f_{\N_u}) &\leq {J}_{\tilde{\N}}(\sum_{i=1}^N \omega_i f_{\N_i}) 
\\&\quad+ \sum_{i=1}^N \omega_{i}\left( \frac{1}{2} {d}_{\mathcal{H} \Delta \mathcal{H}}\left(\N_i, \N_u\right) + \epsilon_i\right)
\\&\quad+4 \sqrt{\frac{2 d \log \left(2NM\right)+\log \left({4}/{\delta}\right)}{NM}}.
\end{aligned}
\end{equation}
\textbf{Extend to multiple unseen clients:}
Given $N_u$ unseen clients, let $\N_j$ be the distribution for the $j$-th unseen client, $j = 1, \cdots, N_u$, and let $\tilde{\N}_u$ be the mixture of unseen clients. Then, the error bound is:
\begin{align*}
    J_{u}(f_{\N_u}) \leq & \sum_{j=1}^{N_u} J_{u_j}(f_{\N_u}) \\
     \leq & N_u{J}_{\tilde{\N}}(\sum_{i=1}^N \omega_i f_{\N_i}) \\
     &+\sum_{j=1}^{N_u}\sum_{i=1}^N \omega_{i}\left( \frac{1}{2} {d}_{\mathcal{H} \Delta \mathcal{H}}\left(\N_i, \N_{u_j}\right) + \epsilon_{ij}\right) \\
     & +4 \sqrt{\frac{2 d \log \left(2 NM\right)+\log \left({4}/{\delta}\right)}{NM}},
\end{align*}
where $\epsilon_{ij}$ is the risk of the optimal hypothesis on the mixture of seen client $\N_{i}$ and unseen client $\N_j$.

\subsection{Corollary~\ref{cor:error_bound_fedbn} of Reduced Error Bound by Using Client-specific BN}
\label{sec:proof_error_bound_interve}
We first give the formal statement of the new error bound with reduced divergence term using our proposed client-specific BN.
\begin{corollary}[generalization bound with client-specific BN]
\label{col:error_bound}
Under the assumptions of Theorem~\ref{thm:initial_bound}, the new generalization error for our method ($f^*$) is bounded as:
\vspace{-1mm}
\begin{equation}
\label{app:error_bound_bn}
\begin{aligned}
  J_{u}(f_u^*) & \leq \hat{J}_{\tilde{\N}}(\sum_{i=1}^N \omega_i f_i^*) + \sum_{i=1}^N \omega_i \epsilon_i \\&\quad+4 \sqrt{\frac{2 d \log \left(2 NM\right)+\log \left({4}/{\delta}\right)}{NM}}.
\end{aligned}
\end{equation}
\end{corollary}
The proof is given below, and the key idea is that, by using our client-specific BN, i.e., using the test statistics to normalize features, the divergence of the distributions of the normalized features between seen and unseen clients decreases. More specifically, if the input data are all Gaussian distributed, our method is tend to obtain the standard Gaussian distribution same as feature distributions during training, thus the divergence becomes zero.

As we assume that $\cH$ is a class of two layer neural network with batch normalization, then the input data is transformed as:
\vspace{-1mm}
\[\z:=(V[W \b x])^{-1/2}(W \b x - \E[W \b x]),\] 
where $W \in \R^{d \times m}$ and $m$ is the number of neurons. Since $\b x$ is Gaussian distributed, the transformed layer $\z$ follows $\N$, where $\N$ is the standard Gaussian distribution with mean $\mathbf{0}$ and covariance matrix as identity matrix. For $\forall f \in \cH$, there exists a function $q$ such that $q(\z) = f(\b x;W)$ and for the true labeling function $g$, let $q_{t}(\z) = g(\x;W_{t})$, where $W_{t}$ is the true model parameter. Thus, we have:
\vspace{-1mm}
\begin{align*}
J_{s}(f) &= \E_{\b x \sim \N_i}[\ell(f(\b x),g(\b x))] \\&= \E_{\z \sim \N}[\ell(q(\z), q_t(\z))] = J_{\N}(q).
\end{align*}
For seen clients, the risk becomes $J_{\N}(q)$ instead of $J_{\N_i}(f)$, which means the covaraite shift is eliminated by BN. 

For unseen client with aggregated BN, the input data is transformed as $\z_v:=\Sigma_s^{-1/2}(W \x - \mu_s)$, where $\mu_s$ and $\Sigma_s$ are the mean and variance matrix calculated using the seen clients data. Let $\N_v$ be the distribution for $\z_v$, then the error bound becomes:
\vspace{-2mm}
\begin{equation*}
\begin{aligned}
J_{\N_v}(q_{u}) &\leq {J}_{{\N}}(\sum_{i=1}^N \omega_i q_{i}) + \sum_{i=1}^N \omega_{i}\left( \frac{1}{2} {d}_{\mathcal{H} \Delta \mathcal{H}}\left(\N, \N_v\right) + \epsilon_i \right) \\&\quad+4 \sqrt{\frac{2 d \log \left(2NM\right)+\log \left({4}/{\delta}\right)}{NM}}.
\end{aligned}
\end{equation*}

For unseen client with proposed client-specific BN, the input data is transformed as $\z_I:=\Sigma_t^{-1/2}(W \x - \mu_t)$, where $\mu_t$ and $\Sigma_t$ are the mean and variance matrix calculated using the seen clients data. Then we have $\z_I \sim \N$, and the error bound becomes:
\vspace{-2mm}
\begin{align*}
    J_{\N}(q_{u}) & \leq {J}_{{\N}}(\sum_{i=1}^N \omega_i q_{i}) + \sum_{i=1}^N \omega_{i}\left( \frac{1}{2} {d}_{\mathcal{H} \Delta \mathcal{H}}\left(\N, \N\right) + \epsilon_i\right)\\&\quad+4 \sqrt{\frac{2 d \log \left(2 NM\right)+\log \left({4}/{\delta}\right)}{NM}} \\
    & = {J}_{{\N}}(\sum_{i=1}^N \omega_i q_{i}) + \sum_{i=1}^N \omega_{i} \epsilon_i \\&\quad+4 \sqrt{\frac{2 d \log \left(2 NM\right)+\log \left({4}/{\delta}\right)}{NM}}.
\end{align*}
The client-specific BN reduces the divergence of distributions of the seen and unseen clients.

\section{Causal Analysis and Proofs}
\label{app:causal_analysis_proof}

In this section we show how to achieve the intervention on $F$ via intervening the surrogate variable $S$. The main proof sketch is using calculus of intervention to exchange the effect of intervening $F$ and $S$. 
We start by presenting the challenges of using back-door adjustment in federated scenario. Then we demonstrate the proof our proposition, which provides a more simple and less costly way for federated testing.

\subsection{Challenges of Back-door Adjustment in FL}
\label{app:proof_proposition}
In this section, we first explain why using back-door adjustment in to tackle issues in federated testing is challenging.

To conduct intervention on a certain vertex $X$ in the causal graph via back-door adjustment, the key idea is summing the conditional probability over a set of parent vertices of $X$. Based on~\cite{peters2017elements}, we first define the valid adjustment set as following:
\begin{definition}[Valid adjustment set]
Consider an causal graph over variables $\mathcal{V}$ and let $Y \notin PA_X$ (otherwise we have $P(Y|do(X))=P(Y)$). We call a set $\D \subseteq \mathcal{V}\backslash \{X,Y\}$ a valid adjustment set for the ordered pair $(X,Y)$ if:
\begin{equation}
\label{app:eq:adjustment}
    P(Y | do(X)) = \sum\nolimits_\mathbf{d} P(Y|X,\mathbf{d})P(\mathbf{d}).
\end{equation}
Here, the sum (could also be an integral) is over the range of $\D$, that is, over all values $\mathbf{d}$ that $\D$ can take.
\label{app:def:valid_adjustment}
\end{definition}
From the valid adjustment set definition, we can see that the intervention $P(Y|do(X))$ can be achieved by adjusting over all values $\mathbf{d}$ that $\D$ can take. Before conducting the adjustment as shown in the right hand side of Eq.(\ref{app:eq:adjustment}), we should make sure the left hand side intervention is identifiable, i.e., all variables in the graph should be measured, otherwise, the causal intervention $P(Y|do(X))$ is not computable. We give a theorem based on Chapter 3~\cite{pearl2009causality} to illustrate the identifiabiltiy and the relationships between identifiabiltiy and adjustment as following:
\begin{theorem}[Identifiability]
\label{app:thm:identifiable}
Consider a causal graph over a subset $\mathcal{V}$ of variables are measured, the causal effect $P(Y|do(X))$ is identifiable whenever $\left\{X \cup Y \cup PA_{X}\right\} \subseteq V$, that is, whenever $X$, $Y$, and all parents of variables in $X$ are measured. The expression $P(Y | do(X))$ is then obtained by adjusting for $PA_X$ as in Eq.(\ref{app:eq:adjustment}).

\end{theorem}

From the Definition~\ref{app:def:valid_adjustment} and Theorem~\ref{app:thm:identifiable}, we can summarize that to conduct a valid back-door adjustment for a pair $(X, Y)$, we should satisfy two conditions:
\begin{enumerate}
    \item The causal effect of $P(Y|do(X))$ is \emph{identifiable}
    \item The set of parent variables $PA_X$ for adjustment are \emph{valid}.
\end{enumerate}

In centralized context, there is no strict on data access, all data distributions in $\D$ can be accessed. The intervention can be computed from the observational distribution and the graph structure, i.e., the causal graph is explicitly defined and all \emph{variables} are measured. Therefore, two conditions are satisfied, making the intervention $P(Y|do(X))$ certainly \textit{identifiable} and can be computed via back-door adjustment.

Considering our FL scenario, all variables are measured under the distributed  paradigm, so the causal effect $P(Y|do(X))$ is identifiable. However, when we want to achieve the intervention by adjusting parent variables (i.e., $\D^s_i$) of $X$, since each client data is strictly kept locally, we cannot find a valid adjustment set for $(X,Y)$. The second condition does not hold, making the back-door adjustment failed. To ensure a successful back-door adjustment, the variables $\D_s^i$ need to be shared, which matches the design in some federated domain generalization methods~\cite{liu2021feddg,fedadg}. But these methods may raise the privacy concern and extra communication cost, compared with these methods, our client-specific BN is more simple and less costly to improve the federated testing on external clients.

\subsection{Proof of Proposition~\ref{thm:intervention}}
First, we introduce the d-separation~\cite{pearl2009causality}, which helps to demonstrate the independency between variables in the causal graph. If two variables are independent, there is no causal effects between them, indicating that the intervention on one variable will not have effects on the other one. 

\begin{definition}[d-separation]
Let $A,B,C$ be the three non-intersecting subsets of variables in a directed acyclic graph $\mathcal{G}$. For any path between two variables, a collider is a variable where arrows of the path meet head-to-head. A path from $A$ to $B$ is said to be blocked by $C$ if either a non-collider on the path is in $C$, or there is a collider on the path and neither the collider nor its descendants are in $C$.
If all paths from $A$ to $B$ are blocked, then $A$ is d-separated from $B$ by $C$, denoted by $A \independent B \mid C$.
\end{definition}

Based on the d-separation, we can show different effects of intervention towards dependent or independent variables, so according to~\cite{peters2017elements}, we introduce the rules of intervention calculus as following:
\label{app:proof_intervention}
\begin{theorem}[Rules of intervention calculus]
\label{thm:do_calc}
Given a directed acyclic graph $\mathcal{G}$ and disjoint subsets of variables $X,Y,Z$ and $W$, let $P(\cdot)$ stand for the probability distribution induced by that model, we have the following rules.

\textbf{Rule 1} (Insertion/deletion of observations):
\begin{align*}
P(Y \mid Z, W, do(X))=P(Y \mid W, do(X)),  \\\quad if (Y \independent Z \mid X, W)_{\mathcal{G}_{\overline{X}}}
\end{align*}

\textbf{Rule 2} (Intervention/observation exchange):
\begin{align*}
P({Y} \mid  do(Z), W, do(X))=P({Y} \mid Z, {W}, do(X)), \\\quad if (Y \independent Z \mid X, W)_{\mathcal{G}_{\overline{X} \underline{Z}}}
\end{align*}

\textbf{Rule 3} (Insertion/deletion of interventions):
\begin{align*}
P({Y} \mid do(Z), {W} do(X))=P(Y \mid W, do(X)), \\\quad if (Y \independent Z \mid X, W)_{\mathcal{G}_{\overline{X}, \overline{Z}(W)}}
\end{align*}
\end{theorem}

Start from our formed causal graph for federated testing, the test-specific statistics $S$ (i.e., mean and variance of batch data in testset) is introduced and can be served as surrogate variable with following conditions satisfied:
(1) $F$ intercepts all directed paths from $S$ to $Y$; and
(2) $P(Y|do(S))$ is identifiable (according to Theorem \ref{app:thm:identifiable}). 
The first condition makes sure the causal effect between $S$ and $Y$ is indirect, i.e., $S$ affects $Y$ totally through $F$, otherwise, the effects of intervening $S$ and $F$ will both be passed to $Y$ directly, we can not eliminate either one intervention. The second condition ensures the intervention $P(Y|do(S))$ can be computed over our graph.

Firstly, the {\emph{Rule 3}} of Theorem \ref{thm:do_calc} can be applied on original intervention $P(Y|do(F))$, having:
\begin{equation}
\begin{aligned}
\label{eq:ex:action}
P(Y|do(F))=P(Y|do(F),do(S)), \\\quad because (Y \independent S \mid F)_{\mathcal{G}_{\overline{F} \overline{S}}}    
\end{aligned}
\end{equation}

The Eq.(\ref{eq:ex:action}) demonstrates that, due to the intervention on $F$, variable $F$ is not be affected by its parent variable $S$, so the external intervention $do(S)$ is independent from the label $Y$. With the independency between $S$ and $Y$, we can add $S$ holding any values into the conditional probability, having $P(Y|do(F), do(S))$.

Then, consider relations in $\mathcal{G}_{\underline{F}\overline{S}}$, intervention on $F$ can be removed based on {\emph{Rule 2}}, having:
\begin{equation}
\begin{aligned}
\label{eq:del_action}
P(Y|do(F),do(S)) = P(Y|do(S)), \\\quad because (Y \independent S \mid F)_{\mathcal{G}_{\underline{F} \overline{S}}}    
\end{aligned}
\end{equation}

This equation shows the conditions for an external intervention $do(S)$ to have same effect on $Y$ as the passive observation $S=s$ (here $s$ is the mean and variance of test-specific data), since in $\mathcal{G}_{\underline{F}\overline{S}}$ all paths from $S$ to $Y$ are eliminated.
Combining Eq.(\ref{eq:ex:action}) and Eq.(\ref{eq:del_action}), we finally conduct the intervention on $F$ with the help of surrogate variable $S$,
\begin{equation}
\label{eq:do_via_iv}
P(Y|do(F)) = P(Y|do(S)).
\end{equation}

Combing the above conditions for introducing the surrogate variable $S$ and rules to exchange intervention, we have the proposition as below:
\begin{proposition}
\label{app:thm:intervention} With the causal graph defined for testing, if a variable $S$ served as surrogate variable with following conditions satisfied:
(1)$F$ intercepts all directed paths from $S$ to $Y$;
(2)$P(Y|{do}(S))$ is identifiable;
(3)$(Y \independent S|F)_{\mathcal{G}^{te}_{\overline{S} \overline{F}}}$; and 
(4)$(Y \independent F|S)_{\mathcal{G}^{te}_{\overline{S} \underline{F}}}$, 
then we have
\vspace{-1mm}
\begin{equation}
\label{app:eq:internvention}
    P(Y|do(F)) = P(Y|do(S)).
\end{equation}
\end{proposition}
\clearpage

\end{document}